\documentclass{jov}

\usepackage{amssymb}
\usepackage{amsmath}
\usepackage{subfig}
\usepackage{graphicx}
\usepackage{comment}            
\usepackage{booktabs}           
\usepackage[colorinlistoftodos]{todonotes}
\usepackage{dirtytalk}
\usepackage{csquotes}           
\newtheorem{hyp}{Hypothesis}
\usepackage{epigraph}

\newcommand{\samples}[1]{
  \def\imgsize{0.10}
  \begin{figure}[!t]
    \centering
    \subfloat[Class 1]
    {\fbox{\includegraphics[width=\imgsize\textwidth]{images/dataset_#1/sample_0_0000.png}}
      \fbox{\includegraphics[width=\imgsize\textwidth]{images/dataset_#1/sample_0_0001.png}}}
    \hspace{2mm}
    \subfloat[Class 2]
    {\fbox{\includegraphics[width=\imgsize\textwidth]{images/dataset_#1/sample_1_0000.png}}
      \fbox{\includegraphics[width=\imgsize\textwidth]{images/dataset_#1/sample_1_0001.png}}}
    \hfill
    \caption{Example images for problem #1 of the SVRT dataset by \protect\cite{fleuret2011comparing}}
    \label{fig:prob#1_example}
  \end{figure}
}

\DeclareMathOperator{\f}{f}

\begin{document}

\title{Evaluating the Progress of Deep Learning for \\ Visual Relational
  Concepts}

\abstract{ Convolutional Neural Networks (CNNs) have become the state of the art method for image classification in the last ten years. Despite the fact that they achieve superhuman classification accuracy on many popular datasets, they often perform much worse on more abstract image classification tasks. We will show that these difficult tasks are linked to relational concepts from cognitive psychology and that despite progress over the last few years, such relational reasoning tasks still remain difficult for current neural network architectures.

  We will review deep learning research that is linked to relational concept learning, even if it was not originally presented from this angle. Reviewing the current literature, we will argue that some form of attention will be an important component of future systems to solve relational tasks.

  In addition, we will point out the shortcomings of currently used datasets, and we will recommend steps to make future datasets more relevant for testing systems on relational reasoning.}

\author{Stabinger}{Sebastian}
{Universität Innsbruck}
{Technikerstrasse 21a, 6020 Innsbruck, Austria}
{https://iis.uibk.ac.at}{sebastian@stabinger.name}
\author{Peer}{David}
{Universität Innsbruck}
{Technikerstrasse 21a, 6020 Innsbruck, Austria}
{https://iis.uibk.ac.at}{d.peer@uibk.ac.at}
\author{Piater}{Justus}
{Universität Innsbruck}
{Technikerstrasse 21a, 6020 Innsbruck, Austria}
{https://iis.uibk.ac.at}{Justus.Piater@uibk.ac.at}
\author{Rodr\'iguez-S\'anchez}{Antonio}
{Universität Innsbruck}
{Technikerstrasse 21a, 6020 Innsbruck, Austria}
{https://iis.uibk.ac.at}{Antonio.Rodriguez-Sanchez@uibk.ac.at}


\maketitle

\section{Introduction}
\label{sec:introduction}
Convolutional Neural Networks (CNNs) have become the go-to method for image classification since \cite{krizhevsky2012imagenet} were able to win the ImageNet competition \cite{deng2009imagenet} by a wide margin. Despite the success of CNNs in the field of image classification, there remain some classification problems that seem to be much more challenging for CNNs and other currently available neural network architectures. Examples for such challenging tasks can be found in a subset of the SVRT dataset by \cite{fleuret2011comparing} or in work inspired by Raven's Progressive Matrices \cite{raven1938raven}. In this paper, we will try to convince the reader that tasks which can be categorized as relational concepts are relevant for practical applications and are still difficult to solve for currently used deep learning architectures. In addition, we will point out, that all currently used datasets to test for relational reasoning have one shortcoming or another. Our hypotheses, which we will try to argue for in this work are as follows:

\begin{hyp}[H\ref{hyp:first}] \label{hyp:first} Attentional mechanisms
  will be an important component to successfully and efficiently learn
  relational concepts.
\end{hyp}
\begin{hyp}[H\ref{hyp:shortcomings}] \label{hyp:shortcomings}
  Relational concepts are more difficult to learn for current neural
  network architectures then other concepts.
\end{hyp}

Deep learning \protect\cite{lecun2015deep} has become the workhorse of the machine learning community in the last 10 years. In the form of Convolutional Neural Networks (CNNs), first introduced by \protect\cite{lecun1989backpropagation}, it has been especially successful in solving computer vision tasks.
Deep learning systems are built from networks of artificial ``neurons''. In the end, such a neuron is a weighted sum of its inputs $x$ with an added bias $b$. The weights $w$ of this sum are also said to be the weights of the neuron. A non-linear function $\f$ (called the activation function) is then applied to the resulting sum:
\begin{equation}
  y(x_1,x_2, ..., x_n) = \f\left(\sum_{j=1}^n\left(x_jw_j+b\right)\right)
\end{equation}
These neurons are generally organized in layers and the outputs of neurons of previous layers become the inputs of neurons of later layers. The weights, therefore, connect neurons of the previous layer to the neurons of the next layer. Deep learning is named after the circumstance that modern architectures are, in comparison to previously used artificial neural networks, very deep (i.e. they are constructed of many, sometimes up to multiple hundred, layers)

The output of a neuron, for a given input, can be changed by modifying its weights $w$. The main idea of deep learning systems is to change the weights of all neurons so that a given input will produce an expected output. Changing the weights by hand is not feasible, so training data is utilized to automatically adapt the weights. Data used for training the network consists of input/output pairs (e.g. images and their correct class). The network is fed with the inputs and a loss function is used to compare the output of the network with the known correct output. More specifically, the loss function returns a numeric value that indicates how close the output of the network is to the correct output.

This whole neural network is in essence a giant formula that is differentiable. Single points of non-differentiability generally don't pose a problem in practice and occur in many modern neural network architectures\footnote{A widely used activation function is the ReLU function $f(x)=\max(0,x)$, which is differentiable for all inputs but $0$. In practical implementations, it is common to return, if possible, one of the one-sided derivatives for points that are not fully differentiable. Even though this is mathematically not completely accurate, it works well in practice.}. Therefore, we can also calculate the gradient of this whole system with respect to the weights. The gradient gives us information on how to change the weights so that the output of the loss function will get slightly lower for the currently presented data. Doing this over and over for known data moves the weights of the neural network to values that will give us the correct output for a certain input.

If the training was done successfully and with enough and representative training data, the resulting network will now also generalize to unseen data and will give us the correct output for previously unseen input (e.g. the correct class for a new image that is not in our training data). This of course only works within limits, i.e. if the training data is statistically similar enough to the unseen data so that the network can generalize. \cite{mri-out-of-distribution} could for example show that for MRI images, even different tissue contrast from the training set leads to much poorer generalization performance. The question of why deep neural networks generalize as well as they do in many cases and at which point generalization breaks down are questions that are not yet answered sufficiently and are still a topic of ongoing research (e.g. \cite{rethink-generalization} and \cite{sensitivity} among many others).

\subsection{Concept Learning}
\label{sec:concept-learning}

\epigraph{Concepts are the glue that holds our mental world together.}{\cite{bigbook}}

The decision which group (or class) a stimulus belongs to is usually called \textit{classification} in the field of machine learning. In cognitive psychology, the same task is more widely known as \textit{categorization} and is thought to be facilitated by knowledge in the form of concepts.

The idea of concepts emerged from the observation that humans categorize and group objects and experiences to be able to efficiently navigate the world and transfer knowledge from one concrete physical object to another. E.g. although every object in a grocery store is unique, we might categorize multiple of them into the concept of ``Tomatoes'' and transfer the knowledge we have obtained from past experiences with other objects in the concept ``Tomatoes'', and even information we have read about the concept ``Tomatoes'', to infer a lot of information about other concrete physical objects (i.e. other tomatoes) that we have never seen before. Therefore, the concept ``Tomatoes'' allow us to infer that we probably would or would not like to eat these concrete tomatoes, although we have never tasted them. Being able to form vast networks and hierarchies of robust concepts is what allows humans to successfully navigate even completely foreign environments and situations. The ability to learn such concepts from observation and experience is called \textit{concept learning}. From the point of view of machine learning, concept learning is therefore all about maximal utilization of and generalization across training data. Three broad types of concepts can be differentiated, namely \emph{perceptual}, \emph{associative}, and \emph{relational} concepts\footnote{ \protect\cite{zentall2002categorization} provide a more in-depth overview of these three concept classes and \cite{bigbook} gives a comprehensive overview of concepts in general}:

\textbf{Perceptual} concepts, also known as similarity--based concepts, group stimuli by their physical similarity. The perceptual concept ``tree'' for example can be learned by the fact that most trees look similar (i.e. the statistical distribution of features of one tree are similar to the statistical distribution of features of another tree).

\textbf{Associative} concepts emerge because multiple stimuli are associated with the same event or outcome. Thus, one member of an associative class can be represented by another member of the same class. A human can, for example, associate the written word ``tree'', the picture of a tree, and an actual physical tree, because all these stimuli convey the same abstract meaning (i.e. in many contexts the word ``tree'', the picture of a tree, and an actual physical tree can stand in for each other). This is for example what allows humans to transfer knowledge gained by reading about trees to actual physical trees.

\textbf{Relational} concepts put multiple entities in a relationship to each other. The Same--Different concept is one of the most studied relational concepts. For a human, it is very natural to attach the label ``same'' to objects if they are similar in some property (e.g. height, colour, movement direction, ...). It is essential to differentiate between \emph{perceptual} and \emph{relational} concepts: A cup might be grouped into the perceptual concept ``cup'' because it looks similar to other cups. Given a scene with multiple cups, a subset of these cups might be grouped if they are more similar to each other than they are to the other cups and the relational concept ``similar cups'' might be applied to this group of cups. This information might for example be used to determine that all of those ``similar cups'' probably are able to hold the same amount of liquid, without having to actually test each individual cup. Another kind of relational concept are transitive relations like ``stronger than'', which can be used to infer a strength hierarchy without having to test one's strength against every member of a group. If A is stronger than I, and B is stronger than A, it is very likely that B will also be stronger than myself, so I can prevent possible injury by not even competing against B. So a perceptual concept can apply to a single entity, while a relational concept can only be applied to at least two entities.

The rest of this paper is structured as follows: In the next section ``Related Work'' we will present experimental evidence that many animals are also able to learn the previously mentioned concept classes, give an overview of often used neural network architectures, and provide an overview of how existing research in the area of deep learning relates to these concept classes. The section ``Current Research on Deep Learning for Visual Relational Concepts'' composes the main part of this paper. We will take a closer look at deep learning research on relational concepts, and showcase that deep learning methods still struggle with such tasks for the most part. In cases where deep learning systems seemingly are able to solve relational concept tasks, we will point out possible flaws in the datasets indicating that it is difficult to conclude whether the neural network learned the real underlying relational concept with currently used datasets. The discussion will try to coalesce all the findings into actionable steps for further research.

\section{Related Work}
\label{sec:animal-studies}
The idea of concepts as well as the three classes of \emph{perceptual}, \emph{associative}, and \emph{relational} concepts emerged from an anthropocentric perspective. Therefore, it is not surprising that humans have no difficulty learning all of them. However, there is also sufficient evidence that at least some animals can learn these concepts to some degree. This indicates that the ability to form abstractions, separate from concrete physical objects, is not something that only humans possess and, more importantly, that not only humans can learn.

Regarding \textbf{Perceptual Concepts}, \cite{herrnstein1964complex} did already show that pigeons can be trained to classify images (e.g. into the classes ``person'' and ``non-person'') and also generalize to new, unseen images, indicating that they can learn perceptual concepts. \protect\cite{schrier1987categorization} showed the same for macaque monkeys, \protect\cite{vogels1999categorization} for rhesus monkeys, \protect\cite{vonk2002natural} for gorillas, and \protect\cite{vonk2004levels} for orangutans. This might not be too surprising, since a bird will readily eat a cherry without having first tasted this specific cherry, but it shows that these concepts do not have to be genetically predetermined, but can be learned from experience, even in animals.

\begin{figure}[t]
  \centering
  \includegraphics[width=0.3\textwidth]{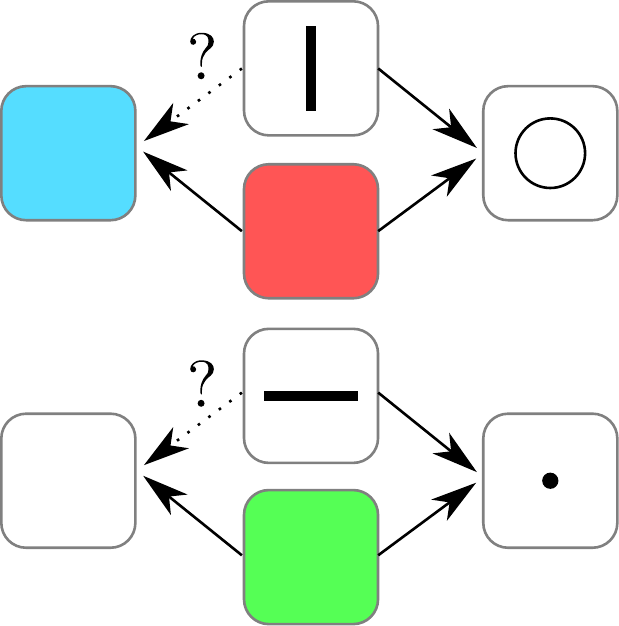}
  \caption{Visualization of a testing procedure employed by \protect\cite{wasserman1992non} to determine whether an animal can learn associative concepts. The animal is trained to select the same response for multiple stimuli (a big circle when shown a vertical line or the colour red and a small dot when shown a horizontal line or the colour green). The colours red and green are later also associated with different responses (a blue light and a white light respectively). The animal is then tested for the remaining two stimuli and the new responses. If the animal did indeed learn associative groups, one would expect that the blue light is selected for a vertical line and the white light is selected for the horizontal line, even though these specific stimuli/response pairs were never presented during training.}
  \label{fig:associative_concept}
\end{figure}

To test whether animals can form \textbf{Associative Concepts}, they can be trained to select the same response for multiple stimuli. An animal can, for example, be trained to respond to the colour red as well as the picture of a vertical line by selecting a big circle. Similarly, a green light and a horizontal line can be associated with a small circle (see \autoref{fig:associative_concept}). The hypothesis is that the red light and vertical line, as well as the green light and the horizontal line, would be grouped in two associative classes because they are linked to the same response. To test whether this hypothesis is correct, the red and green light are later associated with another pair of responses, namely a blue and white light. If associative classes are formed by the animal, testing the vertical and horizontal line as a stimulus and the blue and white light as possible responses should lead to a higher probability of pairing the vertical line with the blue light and the horizontal line with the white light, even though these stimuli/response pairs were never seen by the animal before. \protect\cite{wasserman1992non} performed exactly this experiment and could show that pigeons can learn associative concepts. This might be a more surprising outcome, but it demonstrates that even associating physically completely unrelated stimuli to each other, and therefore being able to transfer knowledge gained from one stimulus to the other, is not something uniquely human. From an evolutionary perspective, it makes sense for animals to possess the ability to form associative concepts since it reduces the amount of, potentially fatal, experiences an animal has to have to learn. This ability is brought to perfection in humans who can learn from the experiences of other humans by communicating abstract concepts.

The same/different task, in which stimuli have to be compared for identity or similarity in one form or another, has been the most thoroughly studied \textbf{Relational Concept} in animals. \protect\cite{zentall1976} showed that pigeons could choose a shape that is identical to a previously presented shape and that this ability also transfers to shapes not seen during training. These results for pigeons have been confirmed multiple times by different researchers in the following years (e.g. \protect\cite{blaisdell2005two} and \protect\cite{katz2006same}). The ability to learn the same/different concept has also been shown for bottlenosed dolphins by \protect\cite{mercado2000generalization}, for infant chimpanzees by \protect\cite{oden1990infant}, for African gray parrots by \protect\cite{pepperberg1987acquisition}, for rhesus and capuchin monkeys by \protect\cite{wright2006mechanisms}, for dogs by \protect\cite{byosiere2017relational}, for rats by \protect\cite{wasserman2012same}, for ducklings by \protect\cite{martinho2016ducklings}, and for bees by \protect\cite{giurfa2001concepts}. This allows animals to transfer information about a concrete object to similar objects, and therefore make learning more efficient. In addition, researchers were able to determine that a wide array of animals are able to use transitive relational concepts to efficiently determine social order (e.g. \cite{grosenick2007fish} for fish, \cite{hogue1996coherent} for hens, and \cite{bond2004pinyon} for pinyon jays).

The fact that many animals can learn concepts from all three concept classes (including relational concepts) suggests that this capability is valuable for agents interacting with and learning from the real world.

\subsection{Concepts and Deep Learning}
\label{sec:conc-deep-learn}

The question of which of the three concept classes can be learned with deep learning has not been systematically studied until now. More generally, to the best of our knowledge, the connection between concept classes and deep learning has not yet been made to the extent presented in this work. We think that this novel viewpoint is useful since the concept classes seem to align quite well with how difficult tasks are for feed forward networks. Specifically, tasks that can be seen as learning relational concepts seem to be more difficult to feed forward neural networks than tasks including other concepts.

Although it is rarely presented from this perspective, Convolutional Neural Networks were specifically developed to solve \textbf{perceptual concept learning}. The architecture of CNNs is specifically designed to classify images using statistical correlations between image patterns of a more and more abstract nature as the information flows to higher layers \cite{cammarata2020thread:}. The tasks for which CNNs are most widely used (i.e. classifying novel images that were not seen during training) are almost identical to the experiments used to show the ability of perceptual concept learning in animals.

One widely used dataset for classification in deep learning research is the one employed in the ImageNet Large Scale Visual Recognition Challenge \cite{deng2009imagenet}, consisting of $1.2$ million training images, categorized into 1000 classes. The Top-5 error rate\footnote{For the Top-5 error rate, five predictions of the correct class are made, and an image is classified correctly if one of the five predictions is the correct one} of humans on this dataset is $5.1\%$, according to \cite{imagenet}. It should be noted, that this number was obtained by only testing a single subject, but it is the only officially published result for humans. The tested subject describes his experience with the task in \cite{imagenet_human_subject}. According to the author, it is difficult to even get an overview of what 1000 classes are available for selection, and fine-grained classification (there are more than 120 different breeds of dog as separate classes in the dataset) are quite difficult for humans. The CNN architecture presented by \cite{he2015delving} first outperformed the $5.1\%$ error rate of the tested human subject with a Top-5 error rate of $4.94\%$, which has steadily fallen to around $1.2\%$ by 2020 \cite{pham2020meta}. Considering that CNNs perform better than humans on many tasks that are similar to the ones intended to detect perceptual concept learning in animals, it is not unreasonable to assume that perceptual concept learning is the prime example of a task that CNNs are exceptionally good at.

To the best of our knowledge, deep learning has never been explicitly tested on \textbf{associative concept learning} in the way animals are usually tested. \cite{mondragon2017associative} proposed the use of deep learning architectures to model associative concept learning but did not perform any experiments.

Despite the lack of explicit experiments in this area, some research does show that a form of associative concepts emerges implicitly in certain circumstances via so-called multimodal neurons. \cite{halle_berry_neurons} were able to show that a biologically inspired, hierarchical CNN, which utilizes sparse coding, produces neurons that strongly activate for persons, even in costume, and the names of those persons. The authors used the biologically inspired ``Locally Competitive Algorithm'' by \cite{lca} to train the network. Other evidence for multimodal neurons comes from research by \cite{distill_multimodal} on neurons in the CLIP architecture by \cite{clip}, which is simultaneously trained on images and image captions using a variant of SGD. The authors were able to find neurons in the CLIP architecture that, for example, strongly activate for images of spiders, spider webs, spiderman in his costume, but also images that contain the text ``Spider''. These multimodal neurons could therefore be interpreted as implicitly learning something akin to associative concepts.

This would indicate that at least some deep learning architectures can implicitly learn associative concepts in their intermediate representations. A direct investigation instead of purely coincidental evidence will be needed to get a better understanding of how well different deep learning architectures can deal with associative concepts.

\textbf{Relational concepts} are interesting since they are not the kind of problems that deep learning was initially conceived for, but are nonetheless very important from a practical point of view for a wide range of computer vision applications. Imagine a robot, asked to pass the ``large cup''. The visual reasoning system of this robot has to be able first to detect cups in its vicinity using perceptual concepts and then use relational concepts to compare the size of the cups to see which one might be considered the ``large cup''. Understanding relational concepts would also allow the robot to transfer knowledge gained about one of the cups to all similar cups, ensuring that training data is utilized most efficiently. In addition, a robot should be able to learn new relational concepts from interactions with humans. Once natural language interfaces to computer systems become commonplace, it will be essential to understand relational concepts since a sizable part of human communication utilizes relations. Therefore, it is not surprising that a lot of the research into relational concept learning (even under a different name) comes from the field of visual question answering (VQA) \protect\cite{wu2017visual}. For these tasks, a system tries to learn how to answer questions about an image, where the questions are asked in the form of natural text. These tasks, unfortunately, mix pure learning of relational concepts with problems from natural language processing (i.e.\ to understand the question). Therefore, we excluded VQA research from this paper since it mixes two separate problem fields, which makes answering the question of whether a system could learn relational concepts even harder than it already is when concentrating on more abstract classification tasks.

Fortunately, over the last few years, researchers have looked at learning relational concepts from images using more abstract tasks to accurately measure the performance of deep learning methods on relational concept learning while minimizing the influence of other confounding factors. In this paper, we will concentrate on such ``pure'' tasks.

\subsection{Deep Learning Architectures}
\label{sec:deep-learn-arch}

Since some specific neural network architectures were used in multiple works that will be presented in this paper, we will briefly give an overview of how they work and why they might be used in certain circumstances:

\textbf{Convolutional Neural Networks (CNNs)}: When applied to image data, the inputs of a neuron can be organized so that the output of the neuron is equivalent to the application of a filter (e.g. Gabor filters) to a specific image region. The kernel of the filter directly corresponds to the weights of the neuron. Since, in most cases, a filter for one region of an image will be equally helpful for other regions, applying the same ``filter neuron'' for all positions of an image is common. This procedure is equivalent to a convolution between the filter kernel and the image. Hence, layers of such neurons are called convolutional layers and neural networks making use of such layers are called Convolutional Neural Networks. In essence, a CNN is purposefully designed to efficiently process and learn from two-dimensional data and utilize spatial invariance, which is present in many images to a certain degree.

Note that learning the weights means that the kernels of the filters used in a CNN are also learned from the data and are not predetermined. For CNNs, it has been shown that this training procedure leads to the layers extracting features which are then combined into more and more complex features as the information flows towards higher layers. This hierarchical extraction of features has been demonstrated exceptionally well in a series of articles by \cite{cammarata2020thread:}.

\begin{figure}[!t]
  \centering
  \includegraphics[width=0.2\textwidth]{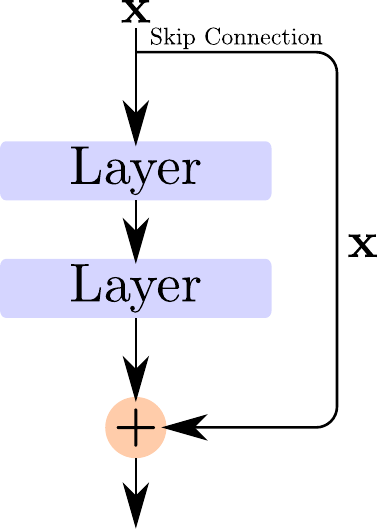}
  \caption{Schematic visualization of a residual block (the main
    building block of Residual Networks)\label{fig:residual}}
\end{figure}

\textbf{Residual Networks (ResNets)}, introduced by \cite{he2016deep}, are one of the standard CNN architectures that are widely used in practice because they overcome one shortcoming of plain CNN architectures: The expressivity of a neural network (i.e. the complexity of the computed function) grows exponentially with the number of layers, but only linearly with the number of trainable parameters. This has been shown theoretically for fully connected networks by \cite{expressivity}, and empirical evidence shows that this likely also holds for CNNs. So deeper networks would generally be preferred to shallower ones. Unfortunately, just stacking more layers leads to the so-called \emph{degradation problem}, where the accuracy a network achieves when being trained on a specific dataset is getting worse the deeper the network gets. This is somewhat counterintuitive since unneeded layers could just be optimized to resemble an identity function, resulting in an output identical to a shallower network. However, this does not happen in practice, indicating that deeper networks are generally harder to optimize if their architecture is not adapted.

Residual Networks mitigate this problem by not only sending an input $\mathbf{x}$ through some of the network layers themselves but by also adding the input to the output of the layers at a later point (see \autoref{fig:residual}). This forwarding of the input to deeper layers is called a shortcut or skip connection. If the layers themselves calculate $\mathcal{F}(\mathbf{x})$ this whole block calculates $\mathcal{F}(\mathbf{x}) + \mathbf{x}$ and is called a \emph{residual block}. By optimizing the weights of the layers, we are optimizing a residual term, hence the name Residual Networks.

For the standard Residual Networks, the layers themselves are convolutional layers, and the whole network consists mainly of a sequence of such residual blocks shown in \autoref{fig:residual}. Although such a residual block should theoretically not be able to learn more than the same network without the skip connection, currently used optimization schemes seem to have a much easier time optimizing this alternative residual rephrasing of the original problem. One reason is that instead of learning an identity function, the layers in a residual block only have to be pushed to output zero since the skip connection already implements the identity function. Another advantage might be that there is always one path for the training signal (via the gradient) to flow to higher layers without going through all the layers themselves.

In \cite{cb-autotune} we were able to present another reason why such skip connections improve the training outcome. We were able to detect layers in neural networks that we named \emph{conflicting layers}, where inputs with different labels collapse to a single point in the activation vector space. We showed theoretically and empirically that conflicting layers degenerate the gradient during training so that weights of the neural network are updated into wrong directions, leading to worse training outcomes. We could show that residual connections skip these conflicting layers.

All these reasons might explain why skip connections seem to perform well in practice and are among the standard architectural components of most modern deep neural networks. Because of this, Residual Networks have become one of the most widely used architectures for computer vision applications.

\begin{figure}[!t]
  \centering
  \includegraphics[width=0.7\textwidth]{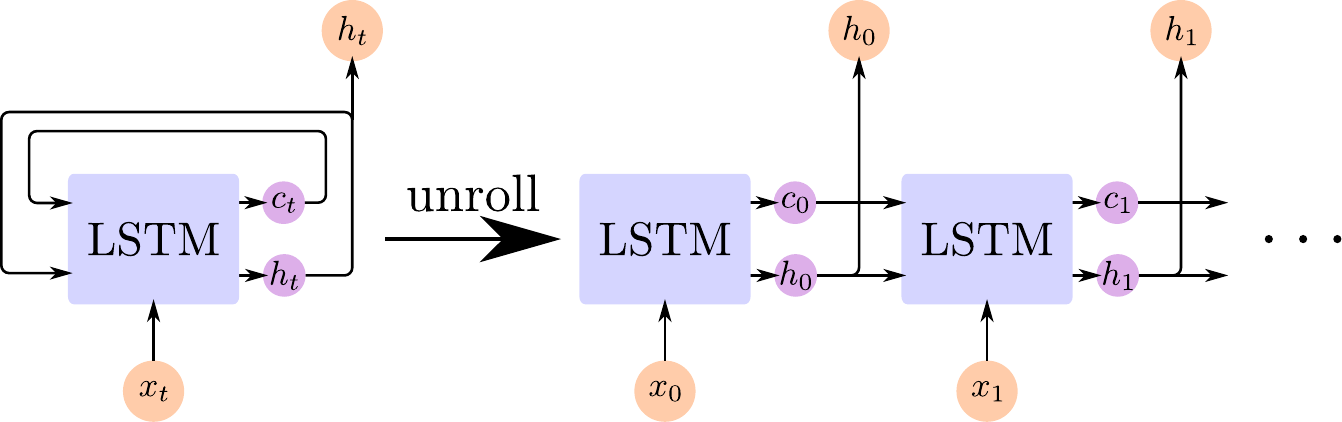}
  \caption{Schematic visualization of how an \emph{LSTM} network is
    being applied to a sequence of inputs. On the left side is the
    general architecture which is applied iteratively to the input
    sequence. The right part demonstrates how this iterative
    architecture can be unrolled to accept a whole sequence at
    once.\label{fig:lstm}}
\end{figure}

\textbf{Long Short Term Memory Networks} (LSTM-networks) are a type of neural network architecture that were developed by \cite{hochreiter1997long} for processing sequences of inputs and are an example of so called recurrent neural networks (RNNs) in comparison to feedforward neural networks like CNNs. Given a sequence $(\mathbf{x_0}, \mathbf{x_1}, \cdots, \mathbf{x_n})$, each vector $\mathbf{x_t}$ of this sequence is iteratively fed to the LSTM as an input, which produces a hidden state $\mathbf{h_t}$ as well as a cell state $\mathbf{c_t}$. $\mathbf{h_t}$ is used as the output of the LSTM for step $t$, but the contents of $\mathbf{h_t}$ and $\mathbf{c_t}$ are also used, together with $\mathbf{x_{t+1}}$, as the input to the LSTM for step $t+1$. The network can therefore forward information to itself in the future, i.e. it can ``remember'' information. \autoref{fig:lstm} shows how such an LSTM-network is applied to a sequence of inputs.

In practice, the LSTM is not iteratively applied to the sequence, but the iterations are unrolled. During unrolling, for a sequence of length $n$, the same LSTM is replicated for each of the $n$ iterations, transforming recurrent connections to feed forward connections, and the resulting bigger system is treated as a single neural network, which can consume the whole sequence at once (see the right side of \autoref{fig:lstm}).

What information is encoded in $\mathbf{c_t}$ and $\mathbf{h_t}$ is not predefined but is learned from the training data by the LSTM via multiple internal neural networks. The unrolled network is trained like any other neural network using a loss function and gradient descent. I.e. an expected output sequence $(\mathbf{y_0}, \mathbf{y_1}, \cdots, \mathbf{y_n})$ is compared to the actual output of the LSTM $(\mathbf{h_0}, \mathbf{h_1}, \cdots, \mathbf{h_n})$ via an appropriate loss function, a gradient with respect to the network weights is calculated, and gradient descent is used to change the weights of the LSTM in the right direction. Note that all copies of the LSTM which were ``produced'' during unrolling are still the same network and have to stay identical also during/after training. Therefore, the weight updates of all instances of the LSTM are aggregated and applied to all instances of the LSTM. Since after unrolling, the gradient propagates through all the duplicates of the LSTM for all the elements of the sequence, the LSTM can ``learn'' to remember some information because it will be helpful later.

Often, the output needed from an LSTM is not a sequence of vectors, but a single vector (e.g. for classifying a sequence), in which case only the output $\mathbf{h_n}$ for the last element in the sequence is compared to an expected output, and all the other hidden states $(\mathbf{h_0}, \mathbf{h_1}, \cdots, \mathbf{h_{n-1}})$ are ignored for the loss.

LSTMs and RNNs, in general, have three advantages over feed forward networks: 1) They can operate over sequences of arbitrary length because the unrolling can be done dynamically. Imagine we want to classify sentences: We can interpret the sentence as a sequence of symbols that we can feed to an LSTM and use the final output of the LSTM to classify some property of the sentence (e.g. its sentiment). Since we can unroll the LSTM to any length we want, we are not restricted by the length of the sentence. At least not in theory; in practice, using an LSTM for much shorter/longer sequences than it was trained on might lead to diminished performance. 2) The fact that the same neural networks process each element of the sequence in the LSTM means that the network can generalize across positions in the sequence (like a CNN can generalize across positions on the two dimensions of an image). For example, if we have to put different panels from a Raven's Progressive Matrix (RPM) test (see \autoref{fig:rpm}) into relation to each other, it is intuitive that features extracted for the upper left panel are probably also going to be helpful for the lower right panel, etc. 3) Through the structure of a sequence we implicitly model that all elements of the sequence are closely related to each other (e.g. all symbols of a sentence, or all panels from an RPM in our case) and most of the relevant information can be inferred by putting them in relation to each other (e.g. the individual symbols in a sentence only really become informative, once they are seen as words etc.), which is helpful if we want to learn relational concepts. One problem with LSTMs when modelling relational concepts is that the entities to be put into relation with each other already have to be separated to feed them into the LSTM as a sequence. This splitting does work for many synthetic datasets, but for real images, the entities first have to be separated, which needs some form of attention, supporting hypothesis \ref{hyp:first}.

\begin{figure}[!t]
  \centering
  \includegraphics[width=0.45\textwidth]{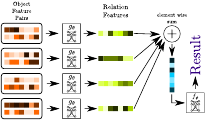}
  \caption{Schematic visualization of a \emph{Relation Network}.
    Features of object pairs are sent through the same neural network
    $g_\theta$ which extracts features encoding the relationship between
    the objects in each pair. These relation features are added to
    accumulate the relational information between all object pairs and
    the resulting vector is interpreted by a neural network $f_\phi$ to
    solve a specific task like
    classification.\label{fig:relation_network}}
\end{figure}

\begin{figure}[!t]
  \centering
  \includegraphics[width=0.45\textwidth]{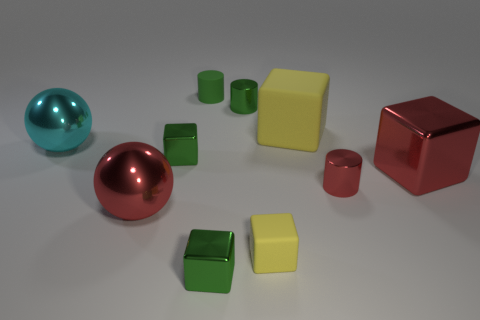}
  \caption{Example image from the CLEVR dataset by
    \protect\cite{johnson2017clevr}. A possible question for this
    image could be: ``What size does the cylinder with the same color
    as one of the spheres have?'', with the correct answer being:
    ``small''.\label{fig:clevr}}
\end{figure}

\textbf{Relation Networks} (see \autoref{fig:relation_network}), introduced by \cite{santoro2017simple}, are based on the principle of applying a neural network $g_\theta$ to all possible ``object''--pairings to detect relationships between them. The big advantage of this is that the application of $g_\theta$ on the object pairs can be done iteratively. Therefore, the network size does not increase with the number of objects to be compared, similarly to how the size of an LSTM does not increase with the length of the sequence to be processed. Objects, in this case, are simply features for which a relationship should be detected. The output of $g_\theta$ for all pairs is added to integrate the information of possible relationships between all object pairs, and the result is sent through an additional neural network $f_\phi$ to produce a final classification. This network architecture was able to achieve superhuman performance on the CLEVR dataset by \cite{johnson2017clevr}, which consists of rendered scenes containing different simple objects of varying sizes, colours, and materials (see \autoref{fig:clevr}). The dataset also includes written questions that, in part, require relational reasoning to be solved (e.g. ``Are there any rubber things that have the same size as the blue metallic sphere?'').

In our opinion, the RN architecture has two main bottlenecks: First, given $n$ objects to be compared, $g_\theta$ has to be evaluated $n \choose 2$ times, so the number of evaluations of $g_\theta$ grows following O($n^2$). If relationships between more than two objects should be handled, the number of needed evaluations proliferates. For relationships between $r$ objects, the network $g_\theta$ has to be evaluated $n \choose r$ times, so the number of evaluations grows with O($n^r$). Therefore, this approach is only practical if the number of ``objects'' can be kept relatively small. Without an attention mechanism, \cite{santoro2017simple} were not able to directly extract features of objects because there was no information about what part of an image is an object. This is the same problem we already mentioned for LSTMs and supports hypothesis \autoref{hyp:first}, which states that some form of attention is an essential component of a system to learn relational concepts. The authors decided to extract features from all positions on a grid over the whole image and handle each position as an object. This method, lacking attention, means that the number of ``objects'' to be compared grows quadratically with the image's resolution. Also, this increase in object pairs results in more and more relation features that have to be integrated, increasing the likelihood that irrelevant relationships between other object pairs wash out helpful information. Secondly, given two object features, the network $g_\theta$ has to recognize the relationship from the information contained in those features alone. If the relationship to be detected is ``similarity'', the representations have to contain all the information to reconstruct the object from it. With more complex objects, these features will become very complex, and a large amount of information must be passed along to $f_\phi$. This bottleneck could be circumvented by iterative processing since the comparison could be made in multiple iterations, and in each iteration, only a tiny part of the whole information from both entities has to be compared.

Although the results of RNs on the CLEVR dataset seem pretty promising, the actual variance encoded in a scene is surprisingly small. There are only 96 different combinations of shape, size, material, and colour. In essence, this means an object in the CLEVR dataset only contains less than 7 bits of relevant information. Some form of positional information, putting the objects in spatial relation to each other, is also needed to solve some of the questions (e.g. ``left of'', ``behind'', ...) contained in the dataset. Still, this will likely not increase the amount of information needed to encode a complete scene by a considerable amount.

Therefore, it is not clear how well the results of RNs on the CLEVR dataset transfer to real-world tasks. Results with different datasets, which will be presented over the rest of this paper, indicate that the performance of RNs decreases for more complex datasets.

\section{Current Research on Deep Learning for Visual Relational Concepts}
\label{sec:deeprelation}
Since most of the research on deep learning is concerned with perceptual concept learning and the systems perform very well on these tasks by design, we will not analyze this group of tasks in more detail. Furthermore, to the best of our knowledge, there is no explicit research on deep associative concept learning, and we will therefore not analyze these tasks in more detail, either. In our opinion, the most interesting tasks can be found within the area of relational concept learning since these tasks seem to be right at the border between solvable and unsolvable tasks for deep learning methods and are also relevant for many practical applications. As mentioned, we will concentrate on ``pure'' tasks from this domain.

\subsection{Work on Ravens Progressive Matrices}
\begin{figure}[tbp]
  \centering
  \includegraphics[width=0.4\textwidth]{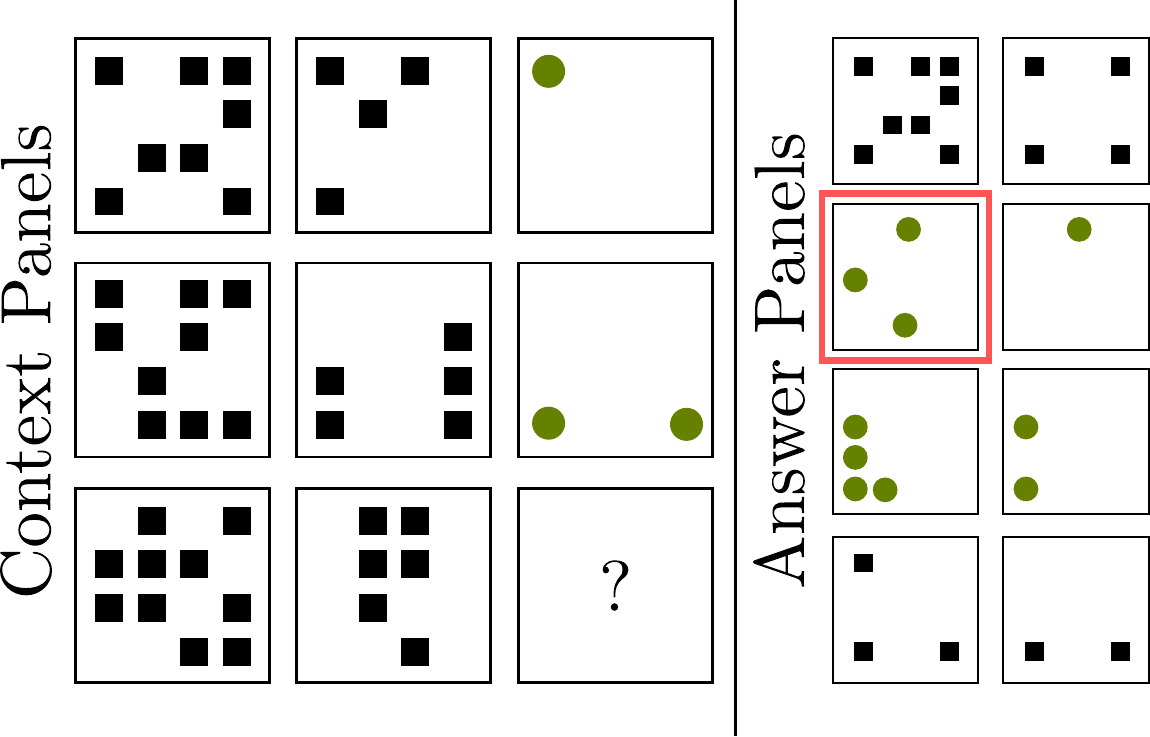}
  \caption{Example of a Raven's Progressive Matrix. The defining
    property of this RPM is that the number of shapes increases by one
    from image to image along the columns while preserving the
    properties of the shapes. The correct solution therefore is the
    second image of the first column. Adapted from
    \protect\cite{santoro2018measuring}~\label{fig:rpm}}
\end{figure}

Raven's Progressive Matrices (RPMs), first presented by \protect\cite{raven1938raven}, are a widely used set of problems to evaluate abstract reasoning and fluid intelligence in humans. Raven's Progressive Matrices consist of a matrix of abstract images related to each other along the columns or rows following specific rules. One of the images is left blank and has to be selected from a set of candidates to relate to the other images following the established rules. Following Occam's razor \cite{schaffer2015not}, the most straightforward rules that can explain the relationships between the images are the correct ones. \autoref{fig:rpm} shows an example of such an RPM.

Learning rules on how a system changes over time and using those rules to predict the future state is, of course, a fundamental property an agent interacting with the world has to master. To learn the rules that drive the change of a system, one has to determine how already observed states relate to each other and RPMs, in essence, test this capability. This is indicative of how important it is to be able to learn relational concepts.

Collections of RPMs used to test humans are not well suited for machine learning since the number of available examples is usually insufficient. Thus, it would not be possible to distinguish the inherent shortcomings of a method from a simple lack of sufficient training data. \protect\cite{wang2015automatic} were the first to use an algorithm to generate an arbitrary number of RPMs. This dataset would have been suited for experiments with machine learning systems, but no such experiments have been performed to our knowledge. Fortunately, multiple datasets have been created by now that follow the basic concept of Raven's Progressive Matrices and are specifically designed for machine learning research.

\subsubsection{Deep Learning and Ravens Progressive Matrices}
\label{sec:deep-learning-ravens}

As far as we can tell, the earliest such work is by \protect\cite{hoshen2017iq} who looked at the performance of neural networks when tasked with choosing or generating the correct continuation of a sequence of changing images, reminiscent of Raven's Progressive Matrices. The networks had to either choose from a predefined set of images (\emph{multiple-choice} task) or had to generate the next image in the sequence directly (\emph{open question} task). Different transformations (e.g. rotation, size, reflection, colour) were used to generate the image sequences.

\begin{figure}[tbp]
  \centering
  \includegraphics[width=0.4\textwidth]{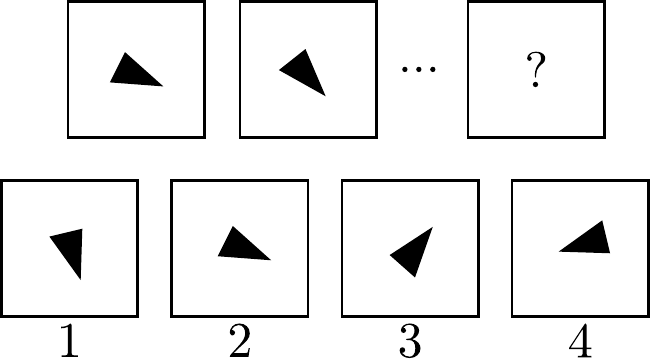}
  \caption{An example for the kind of problems used in the multiple
    choice task by \protect\cite{hoshen2017iq}. The first two images
    are given, showing a triangle that is rotated by a constant angle
    between the first and second image. Four possible continuations of
    this sequence are given, with option 1 being the correct one in
    this case.\label{fig:hoshen}}
\end{figure}

For the \emph{multiple choice} part, a sequence of images is presented to the neural network, together with a set of possible candidates for the next image in the sequence. The network's task is to select the image that continues the underlying pattern. \autoref{fig:hoshen} shows one example of the multiple-choice task.
It was solved by the authors using a network architecture similar to AlexNet (a conventional CNN architecture without skip connections) by \protect\cite{krizhevsky2012imagenet} which was used without pre-training on another dataset first. The image sequence, and possible solution images, were presented to the network as a stack of separate images. Thus, the system did not have to detect and separate the entities and possible solutions independently. The system was able to solve this task with an average accuracy of 97\%.

For the \emph{open question} part, the network did not select an image from a set of possible solutions but generated the next image directly. The network architecture for these problems was based on the DC-GAN architecture by \protect\cite{radford2015unsupervised}, which was also used without pre-training. The performance was measured using the mean squared distance between the ground truth image and the generated image, and the results were also checked qualitatively. The network achieved an average mean squared error of $3.96 \cdot 10^{-4}$, and the resulting images looked qualitatively close to the correct solution.

The results show that even simple CNN architectures are surprisingly good at solving these supposedly complex relational reasoning tasks. Unfortunately, since the networks were trained using $100,000$ images and it is hard to judge the actual variability of the dataset, the achieved accuracy could also be the result of memorization by the network.

As previously mentioned, the images were fed to the network as already separated entities, which is \emph{equivalent to an external attention mechanism}. Following hypothesis \ref{hyp:first}, this already removes one of the main difficulties of such a task. \cite{kim2018not} also pointed this out in a different context. In our opinion, the dataset is therefore not suited for judging a system under real-world circumstances, where such a pre-attention mechanism usually is not present.

\subsubsection{Procedurally Generated Matrices}
\label{sec:proc-gener-matr}

\cite{santoro2018measuring} extended on the ideas by \cite{hoshen2017iq}, and replicated Raven's Progressive Matrices (RPMs) more closely. The authors call this dataset the \emph{Procedurally Generated Matrices} (PGM) dataset, which is freely available. \autoref{fig:rpm} shows a visualization of an example from this dataset. Different architectures were trained and tested on the PGM dataset. The data was again provided to the network as an image stack of 16 separate images (the eight context panels and the eight answer panels). The networks had to select the right panel from the provided answer panels. The rules used for generating the RPMs are pretty elaborate, and we would like to refer the reader to the original paper for more information.

Five different network architectures were tested. 1) A \emph{simple CNN}, 2) a more modern CNN architecture utilizing skip connections in the form of a \emph{ResNet-50} by \cite{he2016deep}, 3) an \emph{LSTM} based on a variant by \cite{zaremba2014recurrent} together with a small CNN for feature extraction, 4) a novel adaptation of a Relation Network \cite{santoro2017simple} which the authors named \emph{Wild Relation Network} (WReN) for which multiple Relation Networks work in parallel, and 5) an adaptation of ResNet which the authors named \emph{Wild-ResNet} for which a ResNet-50 is separately evaluated for each answer panel. A second version of the ResNet architecture, which the authors named \emph{Context-blind ResNet} was used to detect unwanted statistical regularities in the dataset. The Context-blind ResNet was only given access to the answer panels and therefore had to rely purely on statistical properties of the answer set to solve the tasks. In essence, the result from the Context-blind ResNet is the baseline accuracy of a system that does not know the question to be answered. All networks were used without pre-training.

\begin{table}[!t]
  \centering
  \caption{Average accuracy of different architectures tested by
    \protect\cite{santoro2018measuring} on the Procedurally Generated
    Matrices dataset. Adapted from the original paper.}
  \begin{tabular}{cc}
    \toprule
    Model & Accuracy\\
    \midrule
    Blind ResNet & 22\%\\
    CNN & 33\%\\
    LSTM & 36\%\\
    ResNet-50 & 42\%\\
    Wild-ResNet & 48\%\\
    WReN & 63\%\\
    \bottomrule
  \end{tabular}
  \label{tab:barrett1}
\end{table}
The average performance for the whole dataset can be seen in \autoref{tab:barrett1}. The results on the PGM dataset are pretty surprising, considering that the same, simple CNN architecture achieved 97\% accuracy for the dataset used by \protect\cite{hoshen2017iq}. The CNN only performs slightly better than the Blind ResNet, which can be seen as the random baseline accuracy, showing that the dataset by Hoshen and Werman might lack in some way. Either the variability is not large enough in relation to the number of training samples used, which might lead to rote memorization by the network, or the dataset contains statistical correlations that can be used for classification. The WReN architecture performs much better with an accuracy of $63\%$ but is still far from perfect. The research by \protect\cite{santoro2018measuring} would indicate that CNNs, as well as recurrent neural networks, seem to have difficulty with tasks that require more complex relational reasoning, even if the entities are pre-attended. Similarly to the previous dataset, the fact that the entities between which the relations should be detected are already separated makes it difficult to judge how the results on the PGM dataset would transfer to a real-world scenario. Especially, the best performing architecture, utilizing Relation Networks, heavily relies on this pre-attention, supporting our hypothesis \ref{hyp:first} that attention is vital to solve relation reasoning tasks.

\subsubsection{Visual Progressive Matrices}
\label{sec:vprom}

\begin{figure}[t]
  \centering
  \includegraphics[width=0.4\textwidth]{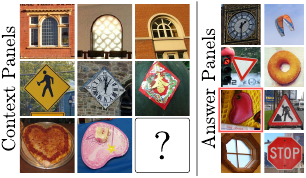}
  \caption{Example from the V-PROM dataset by
    \protect\cite{teney2019v}. In this example, the images in the
    Context Panels are related to each other along the rows by their
    shape. One image is left blank and the correct image, the heart
    shape, has to be selected from the Answer Panels. Adapted from the
    original paper.\label{fig:vprom}}
\end{figure}

\protect\cite{teney2019v} released a conceptually similar dataset named Visual Progressive Matrices (V-PROM) using natural instead of synthetically generated images. See \autoref{fig:vprom} for an example. The authors also include a wide variety of carefully crafted training/testing splits of the dataset to evaluate the generalizability of systems for specific concepts. There are, among others, sets to test how well a system generalizes the concept of counting to unseen numbers and to test if the system generalizes to unseen object categories. The dataset was tested on different network architectures. First, features were extracted from the images using one of two pre-trained CNNs (either a ResNet101 by ~\cite{he2016deep} or a Bottom-Up Attention Network by~\cite{anderson2018bottom}). These extracted features were then interpreted by either a simple multilayer perceptron, a recurrent neural network using gated recurrent units by~\cite{cho2014learning} (a simplified version of an LSTM), the current top-performing method for visual question answering~\protect\cite{teney2018tips}, and a Relation Network by \protect\cite{santoro2017simple}. All systems were trained to either select the correct image or explicitly classify the relationships underlying the images as an auxiliary loss. This auxiliary loss was only used during training to guide the networks to learn good internal representations and proved very useful. In some sense, this loss can be seen as giving additional information to the system about what task it is currently learning. The authors showed that even the best-tested system (again a Relation network) was not able to approach human performance (see \autoref{tab:teney1}).

\begin{table}[!t]
  \renewcommand{\arraystretch}{1}
  \centering
  \caption{Accuracy over the whole V-PROM Dataset from
    \protect\cite{teney2019v}. Adapted from the original paper.}
  \begin{tabular}{ccccc}
    \toprule
    & ResNet & ResNet & B.-up & B.-up\\
    &  & + aux.loss &  & + aux.loss\\
    \midrule
    \textbf{Human accuracy} & 78\% & 78\% & 78\% &78\% \\
    \textbf{RN, shuffled input (base accuracy)} & 13\% & 13\% & 13\% & 13\%\\
    \midrule
    MLP & 41\% & 45\% & 50\% & 56\%\\
    GRU & 43\% & 48\% & 46\% & 53\%\\
    Top-VQA & 37\% & 40\% & 38\% & 41\%\\
    Relation Network (RN) & \textbf{51\%} & \textbf{56\%} & \textbf{55\%} & \textbf{61\%}\\
    \bottomrule
  \end{tabular}
  \label{tab:teney1}
\end
{table}

Similar to the datasets by \cite{hoshen2017iq} and \cite{santoro2018measuring}, the images of the V-PROM dataset were provided to the tested system in an already pre-attended, separated way. This again makes it difficult to judge how well the experimental results would transfer to the real world. Considering that Relation Networks, again the best performing architecture, profit highly from this pre-attended form of data, again strengthens hypothesis \ref{hyp:first}, that an attentional mechanism will be an essential component in a system that can solve relational reasoning tasks.

\subsection{The SVRT Dataset}
\label{sec:25y}
The SVRT dataset \protect\cite{fleuret2011comparing} was created to test the abstract reasoning capability of computer vision systems and compare it to human performance. The dataset consists of 23 problems that are trained for and tested independently. The goal for all the problems is to categorize images (showing abstract shapes) into one of two classes that are separated by some abstract property. For example: In problem 1 (see \autoref{fig:prob1_example} for example images) two shapes are present. For class one, the shapes are different, and for class two, they are identical. Being able to detect similarity is an essential task for any intelligent system to perform searches and identify out of place/novel objects. The SVRT dataset relies heavily on the ability to detect similarity for many of the problems.

\samples{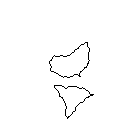}

When the SVRT dataset was created, deep learning was not yet mainstream, so the authors did not test the dataset on those methods. The best performing method tested by Fleuret et al. was Adaboost by \protect\cite{freund1997decision}, using the feature group 3, which includes the \blockquote{\textelp{} number of black pixels in a rectangular subregion of the image for a large number of such regions \textelp*{,} information about the distribution of edges \textelp*{,} spectral properties of the image (Fourier and wavelet coefficients)}~\protect\cite{fleuret2011comparing}. Using a Fourier transform was also recently used to solve the SVRT dataset by \cite{bohn2019few}.

\def\imgsize{0.10}
\begin{figure}[!t]
  \centering
  \fbox{\includegraphics[width=\imgsize\textwidth]{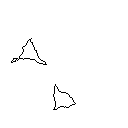}}
  \fbox{\includegraphics[width=\imgsize\textwidth]{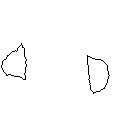}}
  \fbox{\includegraphics[width=\imgsize\textwidth]{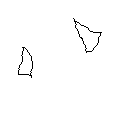}}
  \fbox{\includegraphics[width=\imgsize\textwidth]{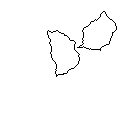}}
  \caption{Examples of incorrectly classified images of the
    ``different-class'' from problem 1 of the SVRT dataset. A
    ResNet-50 network was trained on 28,000 images and the presented
    images were misclassified.}
  \label{fig:svrt_misclassifications}
\end{figure}


Before delving deeper into research done on the SVRT dataset, we would like to mention one potential flaw this dataset might have, in our opinion. A random process is used to generate the shapes for the images. If identical shapes are required, one randomly generated shape is copied pixel by pixel and re-scaled and rotated if necessary. If different shapes are needed, the random shape generation process is used multiple times. This way of producing images means that shapes that should be considered identical are identical (up to scaling and rotation for some of the problems), and shapes that should be considered different are most likely not even roughly resembling each other (see \autoref{fig:prob1_example}). Thus, in most cases, two shapes that approximately resemble each other will be identical. Therefore, it might be enough for a system to detect a few rough local features to be reasonably sure that two shapes are identical without really comparing them. Looking at images misclassified by a well-performing neural network, it is easy to imagine that a network that uses local features to detect similarity might think the presented shapes are identical. See \autoref{fig:svrt_misclassifications} for images that are misclassified by a ResNet-50 architecture with above 90\% accuracy on problem 1, where the goal is to detect whether two shapes in an image are identical or not. For example, the first image shows two shapes with three sharp corners, the second two sharp corners and one bigger ``arch'', etc. In addition, the relations to be learned in the SVRT dataset are relatively simple and mainly consist of recognizing similarity and the spatial orientation shapes. The performance on the SVRT dataset might therefore overestimate a system's ability to learn relational concepts truly.

We started to evaluate deep learning methods on the SVRT dataset in \protect\cite{stabinger2016learning} and greatly extended those experiments in \protect\cite{stabinger201625} by testing how well an old (LeNet by \protect\cite{lecun1989backpropagation}) and a new (GoogLeNet by \protect\cite{szegedy2014going}) CNN architecture performed on the SVRT dataset. We trained LeNet and GoogLeNet for each problem, except for problems 3, 11, and 13, for which we could not generate images with the required size. The models were trained separately for each problem with $40,000$ images and tested using $20,000$ images. The LeNet architecture was trained from scratch, and the GoogLeNet architecture was pre-trained on ImageNet. Even though the presented images look very different from natural images, we found that using a pre-trained network on natural images converges faster during training.

One other goal was to compare the performance of CNNs to that of humans who were tested by \cite{fleuret2011comparing}. Unfortunately, this is not directly possible since individual subjects in practice either achieve 100\% accuracy on a problem if they can figure out the underlying rule separating the classes or achieve accuracy close to chance if they are not able to figure out the rule. Therefore, we report the mean accuracy the group of all tested subjects would be able to achieve.

\autoref{tab:results} shows the results for both tested network architectures, in addition to all other results on this dataset by research presented in this paper at a later point.
\begin{table*}[!t]
  \footnotesize
  \centering
  \caption{Aggregation of results for the SVRT dataset by
    \protect\cite{fleuret2011comparing}. The two groups indicate
    problems which entail Same--Different relations or not.
    [1]~\protect\cite{stabinger201625}, [2] Results of the best
    performing CNNs per problem by \protect\cite{ricci2018same}
    (reconstructed from the published graph) [3]
    \protect\cite{messina2019testing}, [4]~Results with $100$ / $1000$
    / $28,000$ training images by \protect\cite{funke2020notorious}
    (reconstructed from the published graph), [5] Best results per
    problem by \protect\cite{bohn2019few} , [6] Boosting with feature
    group 3 by \protect\cite{fleuret2011comparing}, [7] Human
    accuracies as estimated in \protect\cite{stabinger201625} using
    experimental data by \protect\cite{fleuret2011comparing}}
  \begin{tabular}{rccccccccl}
    \toprule
    Problem & LeNet[1] & GoogLeNet[1] & small CNNs[2] & CorNet-S[3] & ResNet-50[4] & Fourier[5] & Adaboost[6] & Human[7] & Rule\\
    Parameters & 60,850 & 7 Mio & few 10k (varied) & 106 Mio & 23 Mio & 138 Mio & unknown &  & \\
    \# Images & 20k & 20k & 1 Mio & 400k & 100/1k/28k & 20 & 10k & average 6.3 & \\
    Pre-trained & No & ImageNet & No & ImageNet & ImageNet & ImageNet & No &  & \\
    \midrule
    1 & 57\% & 50\% & 62\% & 100\% & 59\% / 88\% / 100\% & 100\% & 98\% & 98\% & Compare\\
    5 & 54\% & 50\% & 67\% & 97\% & 56\% / 69\% / 99\% & 96\% & 87\% & 90\% & Compare \& grouping\\
    6 & 76\% & 86\% & 86\% &  & 58\% / 71\% / 99\% & 51\% & 76\% & 70\% & Compare \& grouping\\
    7 & 53\% & 50\% & 57\% &  & 56\% / 61\% / 100\% & 61\% & 76\% & 90\% & Compare \& grouping\\
    15 & 52\% & 50\% & 68\% &  & 86\% / 100\% / 100\% & 100\% & 100\% & 95\% & Compare\\
    16 & 98\% & 50\% & 76\% &  & 84\% / 100\% / 100\% & 99\% & 100\% & 78\% & Compare\\
    17 & 75\% & 95\% & 88\% &  & 69\% / 83\% / 97\% & 53\% & 67\% & 78\% & Compare \& position\\
    19 & 51\% & 50\% & 60\% &  & 54\% / 76\% / 99\% & 57\% & 61\% & 98\% & Compare\\
    20 & 55\% & 50\% & 56\% & 95\% & 52\% / 56\% / 93\% & 56\% & 70\% & 98\% & Compare\\
    21 & 51\% & 51\% & 59\% & 96\% & 51\% / 70\% / 99\% & 51\% & 50\% & 83\% & Compare\\
    22 & 59\% & 50\% & 63\% &  & 70\% / 97\% / 100\% & 98\% & 97\% & 100\% & Compare\\
    \midrule
    2 & 100\% & 100\% & 100\% &  & 100\% / 100\% / 100\% & 78\% & 98\% & 100\% & Position\\
    3 &  &  & 100\% &  & 95\% / 100\% / 100\% & 58\% & 95\% & 100\% & Position\\
    4 & 98\% & 100\% & 100\% &  & 100\% / 100\% / 100\% & 67\% & 93\% & 100\% & Position\\
    8 & 94\% & 91\% & 95\% &  & 92\% / 99\% / 100\% & 83\% & 90\% & 100\% & Position\\
    9 & 93\% & 100\% & 89\% &  & 81\% / 96\% / 96\% & 51\% & 68\% & 93\% & Size \&  position\\
    10 & 99\% & 100\% & 100\% &  & 97\% / 100\% / 100\% & 84\% & 94\% & 98\% & Position\\
    11 &  &  & 100\% &  & 100\% / 100\% / 100\% & 64\% & 96\% & 100\% & Position\\
    12 & 97\% & 100\% & 100\% &  & 94\% / 100\% / 100\% & 57\% & 84\% & 95\% & Size \&  position\\
    13 &  &  & 91\% &  & 63\% / 97\% / 100\% & 70\% & 67\% & 93\% & Position\\
    14 & 90\% & 100\% & 97\% &  & 73\% / 99\% / 100\% & 68\% & 73\% & 98\% & Alignment\\
    18 & 99\% & 99\% & 100\% &  & 92\% / 99\% / 100\% & 54\% & 99\% & 93\% & Grouping\\
    23 & 87\% & 100\% & 94\% &  & 95\% / 100\% / 100\% & 55\% & 75\% & 100\% & Position\\
    \midrule
    Average & 77\% & 76\% & 83\% & 97\% & 77\% / 90\% / 99\% & 70\% & 83\% & 93\% & \\
    \bottomrule
  \end{tabular}
  \label{tab:results}
\end{table*}
A few surprising facts emerge: First, the best method by Fleuret et al. outperforms many of the more modern architectures on average. Second, the more modern GoogLeNet architecture performs slightly worse than the much older and simpler LeNet architecture. Third, and most interestingly, there seems to be a prominent grouping of problems around the concept of shape comparison. The other results will be discussed later in this paper.
\begin{figure}[!t]
  \centering
  \includegraphics[width=0.7\textwidth]{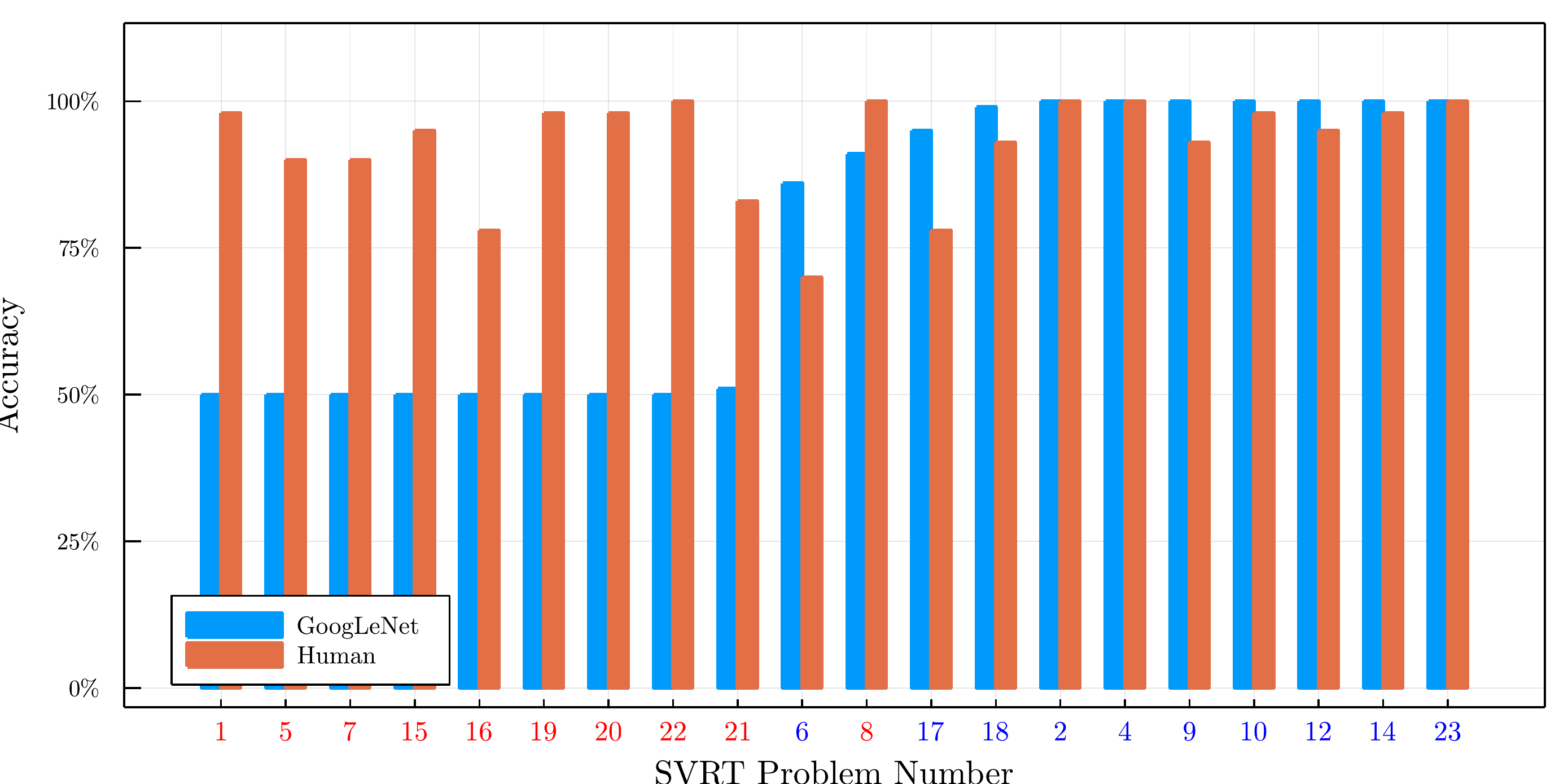}
  \caption{Graphical visualization of the accuracy of GoogLeNet and of
    humans as reported in \protect\cite{stabinger201625} on the
    problems of the SVRT dataset by
    \protect\cite{fleuret2011comparing}.
    \textcolor{red}{Same--Different problems} have a red number, and
    \textcolor{blue}{Spatial--Relation problems} a blue
    one.\label{fig:svrtresults}}
\end{figure}

Problems for which the shapes of the entities are related to each other (\emph{Same--Different} problems) are complex for CNNs and problems where the positions of the entities stand in specific relation to each other (\emph{Spatial--Relation} problems) are easy for CNNs. This is especially evident when looking at a graphical visualization of the achieved accuracies (see \autoref{fig:svrtresults}). For the \emph{Spatial--Relation} problems, LeNet as well as GoogLeNet perform better than the best method used by Fleuret et al. In addition, the newer GoogLeNet performs significantly better than the old LeNet architecture, almost reaching an average accuracy of 100\%. For the \emph{Same--Different} problems, the performance of the CNN architectures is much worse. Both architectures do not achieve an accuracy that is significantly above chance in almost all of the cases.

Three problems seem to go against the general trend (namely problems 6, 16, and 17). A system should theoretically need to perform shape comparison to solve these problems, but we could show that additional information in the dataset enabled the CNNs to correctly classify the images without the need to compare shapes (see \cite{stabinger201625} for a more in-depth explanation). This demonstrates that one has to take great care when creating a dataset to test the performance of machine learning systems since they readily exploit properties of the dataset one did not intend to be used for classification.

\cite{ricci2018same} independently performed very similar experiments to us on the SVRT dataset. The authors also tested convolutional neural network architectures on this dataset but used a whole set of CNNs to check whether the performance difference of Same--Different tasks to Spatial--Relation tasks was influenced by the architecture. We refer the reader to the original paper for a detailed description of the used architectures and training procedures.

\begin{figure}[tbp]
  \centering
  \includegraphics[width=0.7\textwidth]{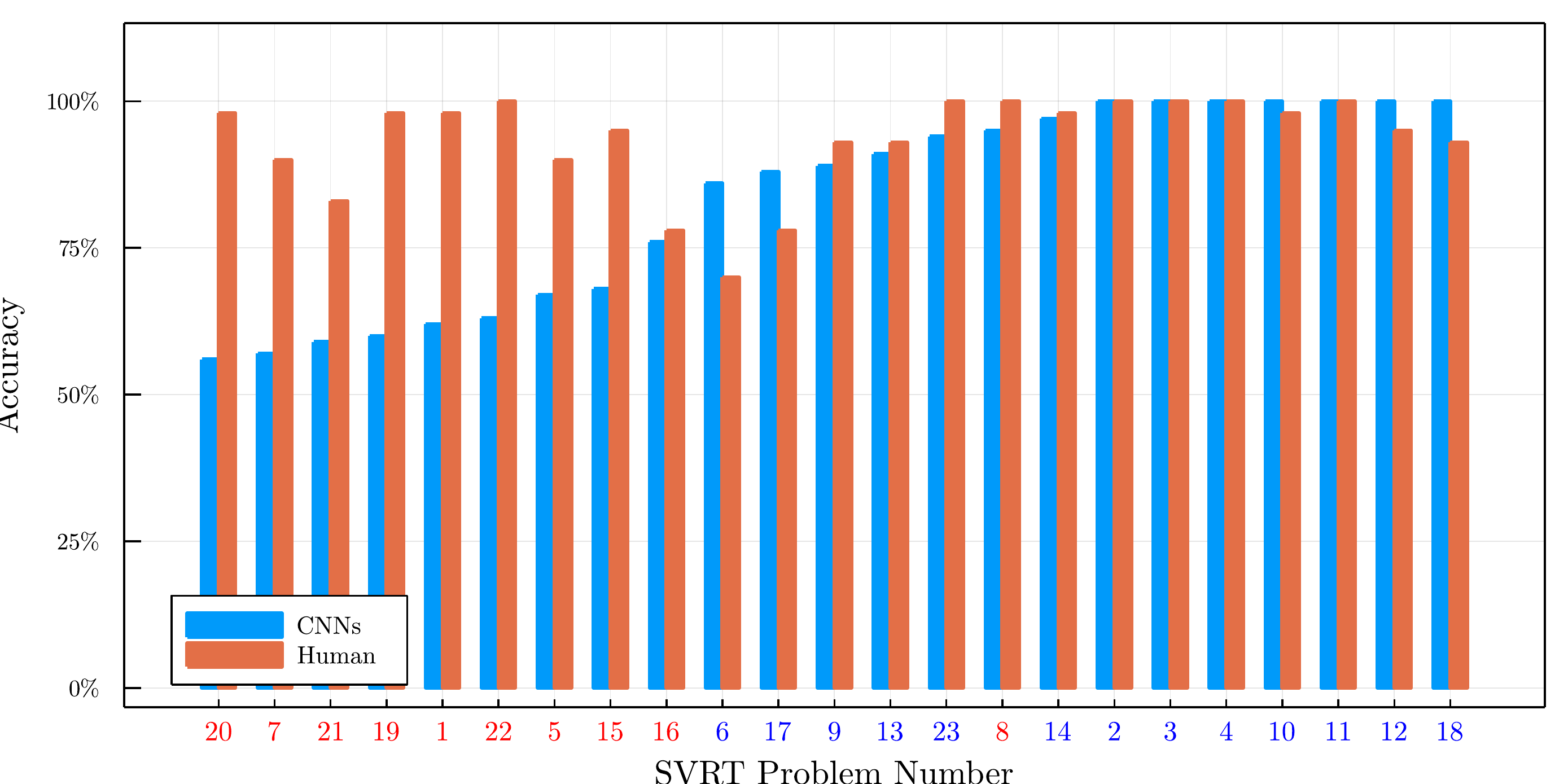}
  \caption{Accuracy achieved by \protect\cite{ricci2018same} and by
    humans in \protect\cite{stabinger201625} for the problems of the SVRT
    dataset by \protect\cite{fleuret2011comparing}.
    \textcolor{red}{Same--Different problems} have a red number, and
    \textcolor{blue}{Spatial--Relation problems} a blue one. Compare
    with \autoref{fig:svrtresults}.\label{fig:serreresults}}
\end{figure}

Ricci et al. confirmed the finding by us that CNNs seem to be particularly challenged by tasks that require the comparison of ``objects''. The authors could also show that the size of the network was less critical for the Spatial--Relation problems of SVRT (i.e. problems where the positioning of shapes is essential) in comparison to the Same--Different problems (i.e. problems that rely on the comparison of shapes). The overall performance reported by \protect\cite{ricci2018same} (see \autoref{fig:serreresults} and \autoref{tab:results}) is higher than what we were able to achieve in \protect\cite{stabinger201625}. This difference in performance is likely a result of using more images for training. \cite{ricci2018same} also put some of the problems in the opposing group, but these differences do not change the overall conclusion.

The main conclusion from these experiments is that convolutional neural networks have greater difficulty detecting Same--Different relations than they do to detect spatial relations. This dichotomy could be explained by spatial relations more closely resembling a perceptual concept since the global arrangement of objects can often be solved by simple pattern matching, whereas Same--Different problems are a classic example of a relational concept. This strengthens our hypothesis \autoref{hyp:shortcomings}, that current neural network architectures have more significant problems with learning relational concepts than learning other concepts.

It is also noteworthy that neither the method presented by \cite{fleuret2011comparing} nor the human experiments show a clear difference between the two groups of problems, so the learning of same--different relations does not seem to be more difficult in general, but especially challenging for convolutional neural networks.

\subsubsection{Solving the SVRT Dataset}
\label{sec:solving-svrt-dataset}

In 2019, \cite{messina2019testing} were able to solve problems 1, 5, 20, and 21 of the SVRT dataset. The authors were able to achieve an accuracy of above 95\% for all four problems using different ResNet architectures by \cite{he2016deep} as well as the biologically inspired CorNet-S architecture by \cite{kubilius2018cornet} (see \autoref{tab:results} for the CorNet-S results), but the authors had to use $400,000$ training images to achieve these results, which might be problematic, considering the discussed potential problems of the SVRT dataset. Both networks were pre-trained on the ImageNet dataset before being fine-tuned for the actual task.

\cite{bohn2019few} were also able to solve many of the Same--Different tasks of the SVRT dataset using deep learning while only utilizing 20 training images (see \autoref{tab:results}). They were able to achieve this by extracting the amplitude spectrum of the Fourier transform of the images. As the authors note, it is well known that peaks in the amplitude spectrum correspond to periodic patterns in the image. The peaks, therefore, encode similarity information in a much easier to use form for machine learning methods. Since the difficult part of the task (finding similarities) was more or less solved in a pre-processing step and not by the neural networks, these results do not change our general conclusions about the performance of neural networks for relational concept learning. Nevertheless, it might be a good idea to add this pre-processing step to systems that have to deal with Same--Different relations in real-world settings. It should be noted that \cite{fleuret2011comparing} already recognized the importance of spectral data since Fourier and wavelet coefficients were already part of the features used in the original SVRT paper and might explain the strong performance of the original method.

\cite{funke2020notorious} were finally able to achieve accuracies above 90\% for all problems of the SVRT dataset (see \autoref{tab:results}) without using special pre-processing steps and with a more reasonable amount of $28,000$ training images, using a ResNet-50 architecture that was pre-trained on ImageNet. Our hypothesis \autoref{hyp:first}, that relational concept learning is more difficult for current neural network architectures than other concepts, is still valid, since the results when training on $100$ and $1000$ images still exhibit a big difference between Same--Different problems (which is a relational concept) and Spatial--Relation problems (which are closer to perceptual concepts). In addition, our experiments in \cite{stabinger2021nonlocal} on a much more tightly controlled Same--Different task (inspired by the PSVRT dataset, presented in a later section) showed that ResNet does not perform well on Same--Different tasks in general. This might indicate that the good results by \cite{funke2020notorious} are more indicative of the previously discussed shortcomings of the SVRT dataset than the suitability of ResNet architectures for solving relational concept learning.

\subsection{The Bongard Problems}
\label{sec:bongard-problems}

\begin{figure}[t]
  \centering
  \includegraphics[width=0.4\textwidth]{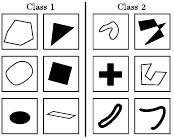}
  \caption{Example of a ``Bongard Problem''. The differentiating
    property in this case is that the images in class 1 show convex
    shapes while the images in class 2 show concave
    objects.\label{fig:bongard}}
\end{figure}

The SVRT dataset is somewhat reminiscent of the problems presented by \cite{bongard1970} as examples of problems a neural network would never be able to solve. It has to be noted that the tasks by Bongard were more difficult than those of the SVRT dataset because the goal was not to classify images but to give a textual description of what separates the two classes. Hofstadter popularized similar problems with his book ``Gödel, Escher, Bach: An Eternal Golden Braid'' \protect\cite{douglas1979godel}. \autoref{fig:bongard} shows an example for such a ``Bongard problem''. The goal is to describe what abstract property separates the images on the left from those on the right. In the case of \autoref{fig:bongard}, the images on the left show convex shapes while the images on the right show concave shapes. To our knowledge, \protect\cite{depeweg2018solving} are the only researchers in recent years that tried to solve Bongard problems as they were originally intended (i.e. trying to generate an explanation of how two sets of abstract images differ), but they did not use deep learning to do so. \protect\cite{nie2020bongard} created a dataset, called Bongard LOGO, which is inspired by the Bongard problems. Unfortunately, the dataset does not contain relational tasks, so we will not cover them in this paper. \protect\cite{yun2020deeper} worked on actual Bongard problems using deep learning as part of their system, but only tested few-shot classification and also not the original task of generating descriptions. Considering the recent success in image caption generation by, e.g. \protect\cite{xu2015show}, \protect\cite{donahue2015long}, \protect\cite{fang2015captions}, and many more, the Bongard problems, in their original form, might be an interesting topic for future deep learning research.

\subsection{The Chess Dataset}
\label{sec:chess-dataset}

As discussed, the SVRT dataset has its flaws. In \cite{stabinger2017evaluation} we tried to create a more robust dataset, which also more closely resembles natural images, by rendering them in a semi--naturalistic way using a 3D rendering software \protect\cite{blender}. Each image shows one or two chessboards with red pawns randomly placed on the field.

The dataset consists of two distinct tasks: The goal of the \emph{identity task}, showing two chessboards in each image, is to detect whether the pawn positions are identical on both boards. The goal for the \emph{symmetry task}, showing one chessboard in the images, is to detect whether the pawn placement is symmetric. The difficulty of both tasks was controlled by allowing translation of the chessboards and the camera, movement of the camera on a virtual half-sphere around the chessboards, and a varying amount of pawns that break the identity/symmetry property. Example images can be seen in \autoref{fig:sim} and \ref{fig:sym}.

\def\thumbwidth{30mm}
\def\imgspace{10mm}
\begin{figure*}[!t]
  \centering
  \subfloat[Fixed camera position]{
    \includegraphics[width=\thumbwidth]{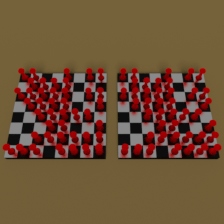}
    \label{fig:simfix}
  }
  \hspace{1mm}
  \subfloat[Random camera translation]{
    \includegraphics[width=\thumbwidth]{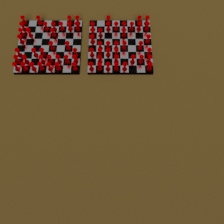}
    \label{fig:simtrans}
  }
  \hspace{1mm}
  \subfloat[Random board position]{
    \includegraphics[width=\thumbwidth]{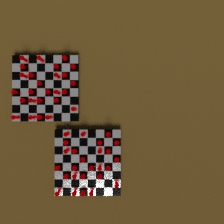}
    \label{fig:simrand}
  }
  \hspace{1mm}
  \subfloat[Random camera position on a sphere]{
    \includegraphics[width=\thumbwidth]{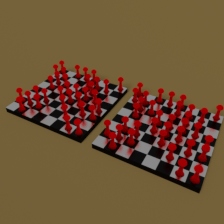}
    \label{fig:simrot}
  }
  \caption{Variations of the identity task in the chess
    dataset.\label{fig:sim}}
\end{figure*}

\begin{figure*}[!t]
  \centering
  \subfloat[Fixed camera position.]{
    \includegraphics[width=\thumbwidth]{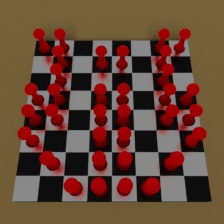}
    \label{fig:symfix}
  }
  \hspace{2mm}
  \subfloat[Random camera translation.]{
    \includegraphics[width=\thumbwidth]{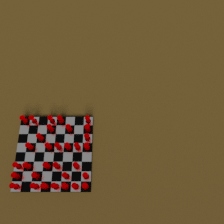}
    \label{fig:symtrans}
  }
  \hspace{2mm}
  \subfloat[Random camera position on sphere.]{
    \includegraphics[width=\thumbwidth]{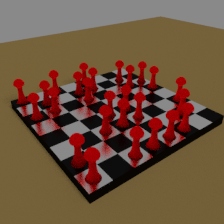}
    \label{fig:symrot}
  }
  \caption{Variations of the symmetry task in the chess
    dataset.\label{fig:sym}}
\end{figure*}
We trained AlexNet by \protect\cite{krizhevsky2012imagenet}, VGG16 by \protect\cite{simonyan2014very}, and GoogLeNet by \protect\cite{szegedy2014going} on all variations of the two tasks with one, five, and ten out of place pawns. All networks were pre-trained on the ImageNet dataset by \cite{deng2009imagenet}, after which the last layer was replaced by a new, randomly initialized layer, to conform to the smaller number of classes of the chess dataset (from 1000 classes down to 2) and the whole network was trained on the chess dataset, without fixing any of the pre-trained layers.

In addition, we had to employ a training scheme that is related to curriculum learning, first proposed by \cite{bengio2009curriculum}. To learn the more complicated variants of a task, we started from networks already successfully trained on easier variations of the same task. Without employing this training scheme, we could not train networks to solve the tasks with only one pawn that breaks symmetry/identity.

The results showed that the symmetry task is considerably more manageable than the identity task, but both tasks can not be learned in all cases. For example, GoogLeNet was not able to achieve accuracies significantly above chance for the most challenging task (i.e. camera rotation with only one out of place pawn) and for the identity task, the network was not able to perform better than chance on any of the tasks with camera rotation, which supports hypothesis \autoref{hyp:shortcomings}.

Still, the results on the identity task were surprisingly good, considering that GoogLeNet is not able to solve the simple task of comparing two shapes from the SVRT dataset and the chess dataset, with random placement of the checkerboards, seems much more complicated, but can be solved quite well by GoogLeNet.

One explanation might be that our curriculum learning approach might be very helpful for such abstract tasks. Unfortunately, it is not immediately clear how to transfer curriculum learning to the SVRT or other datasets, or even more real-world scenarios, because the difficulty of the produced samples can not easily be controlled. In addition, the very regular, never changing, and easy to detect grid of the checkerboard might help the network extract the needed pawn positions, despite the high variability of the images.

\subsection{The Parametric SVRT Dataset}
\label{sec:psvrt-dataset}

\begin{figure}[tbp]
  \centering
  \includegraphics[width=0.30\textwidth]{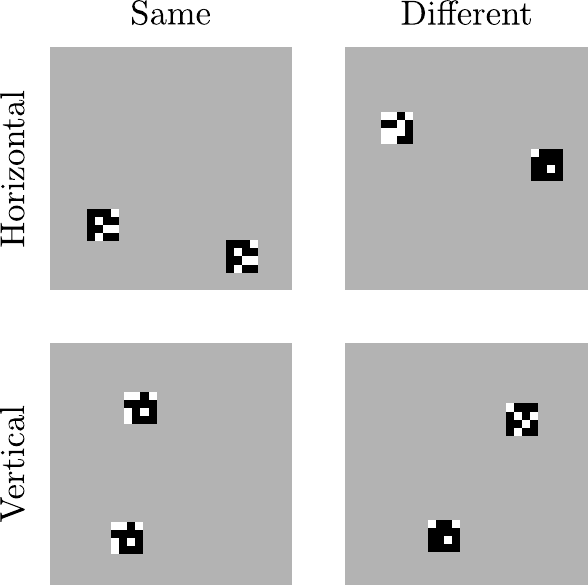}
  \caption{Examples for all four class combinations of the PSVRT
    dataset presented by \protect\cite{ricci2018same}. An image can be
    \emph{same} or \emph{different}, depending on whether the two
    patches show the same pattern and \emph{horizontal} or
    \emph{vertical}, depending on the orientation of the two patches.
    Adapted from the original paper.\label{fig:psvrt}}
\end{figure}

Similar to our reasoning for the chess--dataset, \cite{ricci2018same} recognized that the generation procedures for the SVRT dataset are too unpredictable to lead to reliable conclusions. The authors specifically mention that it is sometimes unclear whether a problem can not be solved because of the relations the network has to learn or because the variability of the images (i.e. the size and number of shapes) has increased. To further investigate the findings from the SVRT dataset that CNNs are better at learning spatial relations than they are at learning Same--Different relations, the authors did a second experiment where they created their own, simple dataset. They call this the \emph{parametric SVRT (PSVRT)} dataset. In the PSVRT dataset, each image contains two patches, composed of black and white boxes, on a neutral background. Examples for this dataset can be seen in \autoref{fig:psvrt}. The two patches have two properties that can be used for classification. The first is the Same--Different relation depending on whether the two patches show the same black and white pattern. The second property is the spatial positioning of the patches, depending on whether the two patches are oriented horizontally or vertically to each other. Three parameters control the amount of variability in the images: The size of the patches, the number of patches, and the image size. Setting up the dataset in this way allows a system to learn the Same--Different as well as the Spatial--Relation problem with identical images and ensures that the image complexity and variability is constant between the two problem sets. We refer the reader to the original paper for further implementation details.

Using PSVRT, the authors were able to show a sharp dichotomy between solving Spatial--Relation and Same--Different tasks as well. The networks consistently learned the classification early in the training procedure for spatial relation tasks and achieved high final accuracy. For Same--Different tasks, the performance was highly dependent on the image size. Bigger image sizes led to slower training, lower-end accuracy and resulted in the networks failing more often at learning the task at all, depending on the random seed used for initialization of the network. In addition, the size of the network (the number of parameters) did not influence the achievable accuracy much for Spatial--Relation tasks but did so for Same--Different tasks. Since the same images were used in both experiments, the authors conclude that image variability was not what hindered CNNs, and learning the Same--Different relation problem itself is what is more difficult for the networks, supporting hypothesis \autoref{hyp:shortcomings} that relational concept learning is more difficult for current neural network architectures than learning other concepts.

The authors hypothesize that the network learns subtraction templates to solve the Same--Different task because the image's patch size and number do not seem to influence the achievable accuracy. The authors argue that more subtraction templates would only be needed if the number of possible patch positions changes, increasing exponentially with growing image size. Unfortunately, the authors do not provide a more detailed explanation for their hypothesis.

The PSVRT dataset has one unfortunate flaw: The patches are not matched for the number of black and white pixels if they are different. Therefore, a simple comparison of the sum of all pixel values between different image parts is sufficient to ``compare'' the patches in many cases. If this is what the authors mean by subtraction templates, then we agree that the networks might use this, but we would argue that this is a flaw of the dataset and not an explanation of how comparison, in general, could be solved by a CNN. Future experiments should test the PSVRT dataset with patches with a matched number of black and white pixels.

\subsubsection{Relation Networks and Siamese Networks Applied to the PSVRT
  Dataset}
\label{sec:relat-netw-clevr}

\cite{kim2018not} extended the work by \cite{ricci2018same} by testing the PSVRT dataset with two additional network architectures. The first was a \emph{Relation Network} (RN) proposed by \cite{santoro2017simple}, which was specifically designed to learn relationships between objects. \cite{kim2018not} hypothesize that the original performance of the architecture on the CLEVR dataset mainly stems from memorization since, as previously mentioned, the dataset only has a minimal amount of variation. The authors could support this hypothesis by testing RNs on the PSVRT dataset and showing that performance decreases with image size in the same way for relational networks as it does for CNNs until the architecture can not learn the task at all at an image size of $180 \times 180$ pixels. As previously argued, we think that the increased number of relation features that have to be integrated might pose another problem. Of course, this could be circumvented using attention, strengthening hypothesis \autoref{hyp:first}. Also, with increasing patch size, it might become challenging to pass all needed information to the network that integrates all relationships (i.e. $f_\phi$).

\begin{figure}[!t]
  \centering
  \includegraphics[width=0.45\textwidth]{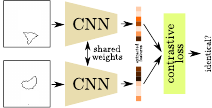}
  \caption{Schematic visualization of a \emph{Siamese Network}. Two
    images to be compared are passed through the same CNN to extract
    high level features. These features are then compared using a
    contrastive loss to determine the similarity of the original
    images.\label{fig:siamese}}
\end{figure}

The second architecture tested by \cite{kim2018not} is a type of \emph{Siamese Network}, first proposed by \cite{bromley1994signature}. Siamese networks were specifically designed to make Same--Different decisions for images. The caveat of Siamese Networks is that the network expects the objects to be compared as separate images (i.e. pre-attended) (see \autoref{fig:siamese} for a schematic visualization). As \cite{kim2018not} point out, this splitting into two images can be interpreted as a kind of attention mechanism simulating the effects of perceptual grouping. The authors were able to show that such Siamese Networks can solve PSVRT successfully, not showing a qualitative performance difference between the Same--Different and the Spatial--Relation task. The performance was also independent of the image parameters (i.e. image size, patch size, and the number of patches). These results are in support of hypothesis \autoref{hyp:first}, which states that attentional mechanisms are an important component in solving relational concept learning. We could also show in \cite{stabinger2021nonlocal} that just separating the entities to be compared into different channels of a normal CNN made a task similar to PSVRT considerably easier.

\section{Discussion and Future Work}
\label{sec:discussion}

In the beginning, we stated two hypotheses: With Hypothesis \autoref{hyp:first} we stated that \emph{``Attentional mechanisms will be an important component to successfully and efficiently learn relational concepts''}. We have shown that Relation Networks, which generally perform very favourably on relational reasoning tasks, need attention to keep the number of object comparison operations low. This is especially important if relations between more than two objects have to be detected. \cite{kim2018not} as well as our research in \cite{stabinger2021nonlocal} shows that pre-attending data considerably improves performance for relation tasks. In addition, as can be seen in \autoref{tab:datasets}, this pre-attention is already an integral part of many datasets that are currently used to test systems for learning relational concepts.This demonstrates that many researchers have recognized this need, even if they did not communicate or consider this themselves from the viewpoint of attention. In our opinion, this explains why the results on these datasets are surprisingly good, considering that the straightforward PSVRT dataset can only be solved using massive amounts of training data, and even in these cases, the results are far from perfect. We think that datasets with a form of pre-attention grossly overestimate the performance, which could be expected under real-world conditions, where such a form of pre-attention is not available.

One promising group of architectures that integrates attention at its core and has gained traction in many fields of deep learning over the last year are transformer architectures. \cite{attention_is_all_you_need} first proposed these for the field of machine translation, \cite{bert} later generalized them to many other natural language processing tasks, and by now they are also heavily researched for many other tasks, including computer vision by \cite{visiontransformer} and \cite{perceiver}, among many others. All transformer architectures contain a self-attention mechanism as one essential building block and, therefore, should be a promising architecture to study for relational reasoning tasks.

With Hypothesis \autoref{hyp:shortcomings} we stated that \emph{``Relational concepts are more difficult to learn for current neural network architectures than other concepts''}. We would argue that despite the progress in recent years, it is still evident that deep learning methods have a weakness in relational reasoning tasks. Results are either not on par with human performance (Procedurally Generated Matrices, V-PROM, PSVRT, Chess Dataset), might be results of weak datasets (IQ Dataset, SVRT, CLEVR), or are unrealistic because the datasets have an attention mechanism embedded in the way the data is presented (IQ Dataset, PGM, V-PROM). Our work~\cite{stabinger201625} was the first to show this divergence in performance for different kinds of concept learning. This dichotomy was later also shown by \cite{ricci2018same} and is currently demonstrated exceptionally well with the PSVRT dataset by \cite{kim2018not}. In our opinion, the recent advances on the SVRT dataset are more indicative of possible shortcomings of the dataset and less of advances of the methods, especially considering the poor performance of the same architectures on the conceptually very similar PSVRT dataset. All of the datasets we presented in this paper have, in our opinion, one or more problems (see \autoref{tab:datasets}).

We think the PSVRT dataset by \cite{ricci2018not}, with an added restriction to pixel value matched patches, to prevent the system of using a simple sum to compare patches, would likely provide the cleanest datasets to test relational reasoning capabilities while minimizing the chance of introducing secondary features a deep learning system can use to ``cheat'' at the task.

\begin{table}[t]
  \centering
  \caption{Datasets presented in this paper and possible problems with
    them}
  \begin{tabular}{rll}
    \toprule
    Name & Citation & Problems\\
    \midrule
    IQ Dataset & \cite{hoshen2017iq} & Dataset is pre-attended, unknown variance\\
    Procedurally Generated Matrices & \cite{santoro2018measuring} & Dataset is pre-attended\\
    V-PROM & \cite{teney2019v} & Dataset is pre-attended\\
    SVRT & \cite{fleuret2011comparing} & Possibly low variance of the images\\
    PSVRT & \cite{kim2018not} & Can be solved using pixel value sums\\
    CLEVR & \cite{johnson2017clevr} & Low variance of the scenes\\
    Chess Dataset & \cite{stabinger2017evaluation} & Static checkerboard might give too many hints\\
    \bottomrule
  \end{tabular}
  \label{tab:datasets}
\end
{table}

In our opinion, even simple bottom-up attention will not be sufficient to solve relational tasks efficiently and iterative attention shifts will be necessary to efficiently solve many relational concept learning tasks in the real world. Attention solves the problem of separating entities to be compared, but does not solve the problem of information density. For example, if two objects have to be compared for identity, all information about the two objects has to be forwarded to a subsystem that can decide on identity. As the objects variability increases, this will likely mean that the layers transporting this information, and the network deciding on identity, will both grow rapidly, making the system inefficient and data-hungry. We theorize that iteratively shifting attention will more favourably balance network size with computation time. We also think that the necessarily shared parameters and substructures will lead to a reduced need for training data and better generalization for relational tasks. This is already partly put into effect in Relation Networks by the iterative application of $g_\theta$, which might be one of the reasons it performs better on average for relational tasks than most other network architectures.

We think future research should concentrate on creating datasets that test for relational reasoning, without providing a form of pre-attention or introducing unwanted features that can be used to ``cheat'' the task. Creating non pre-attended instances of existing datasets (like PGM, V-PROM and the IQ Dataset) might be able to further demonstrate the importance of attention for such tasks. Architectures with iterative processing of the input and a mechanism to shift attentional focus between the iterations are currently not very popular, but should be investigated more deeply with respect to relational concepts.


\section{Conclusion}
\label{sec:conclusion}
In this study, we have summarised and interpreted current deep learning research from the perspective of concept learning. We were able to show that \emph{perceptual concepts} are easily solved by deep learning methods since they were initially developed for this class of problems. \emph{Associative concepts}, even though preliminary evidence suggests that at least some deep learning architectures do implicitly learn such concepts in their intermediary representations, seem to have not been studied until now in the field of deep learning. Thus, we focused our analysis on work that can be classified as learning visual \emph{relational concepts}.

We hope we were able to convince the reader that relational concepts are of practical importance and seem to be particularly difficult for current neural network architectures to learn. We also demonstrated that attentional mechanisms would be, together with a form of an iterative attentional shift, an essential component in solving these problems in the future.

We have also demonstrated that many of the currently used datasets are not ideal and likely overestimate the actual performance of tested systems. Many datasets have a form of pre-attention built-in, or the complexity and variability of the produced samples might be overestimated. New datasets which take these findings into account will therefore have to be created for future research.

\bibliographystyle{jovcite}
\bibliography{refs.bib}

\begin{thebibliography}{}

\bibitem[\protect\citeauthoryear{%
Anderson%
\ \protect\BOthers{.}}{%
Anderson%
\ \protect\BOthers{.}}{%
{\protect\APACyear{2018}}%
}]{%
anderson2018bottom}%
\APACinsertmetastar{%
anderson2018bottom}%
Anderson, P.%
, He, X.%
, Buehler, C.%
, Teney, D.%
, Johnson, M.%
, Gould, S.%
\BCBL{}\ \BOthersPeriod{.}%
\unskip\
\newblock
\APACrefYearMonthDay{2018}{}{}.
\newblock
\BBOQ{}\APACrefatitle{Bottom-up and top-down attention for image captioning and
  visual question answering}{Bottom-up and top-down attention for image
  captioning and visual question answering}.\BBCQ{}
\newblock
\BIn{} \APACrefbtitle{Proceedings of the IEEE Conference on Computer Vision and
  Pattern Recognition}{Proceedings of the ieee conference on computer vision
  and pattern recognition}\ (\BPGS\ 6077--6086).
\PrintBackRefs{\CurrentBib}

\bibitem[\protect\citeauthoryear{%
Barrett%
, Santoro%
, Hill%
, Morcos%
\BCBL{}\ \BBA{} Lillicrap%
}{%
Barrett%
\ \protect\BOthers{.}}{%
{\protect\APACyear{2018}}%
}]{%
santoro2018measuring}%
\APACinsertmetastar{%
santoro2018measuring}%
Barrett, D.%
, Santoro, A.%
, Hill, F.%
, Morcos, A.%
\BCBL{}\ \BBA{} Lillicrap, T.%
%
\unskip\
\newblock
\APACrefYearMonthDay{2018}{}{}.
\newblock
\BBOQ{}\APACrefatitle{Measuring abstract reasoning in neural
  networks}{Measuring abstract reasoning in neural networks}.\BBCQ{}
\newblock
\BIn{} \APACrefbtitle{International Conference on Machine
  Learning}{International conference on machine learning}\ (\BPGS\ 4477--4486).
\PrintBackRefs{\CurrentBib}

\bibitem[\protect\citeauthoryear{%
Bengio%
, Louradour%
, Collobert%
\BCBL{}\ \BBA{} Weston%
}{%
Bengio%
\ \protect\BOthers{.}}{%
{\protect\APACyear{2009}}%
}]{%
bengio2009curriculum}%
\APACinsertmetastar{%
bengio2009curriculum}%
Bengio, Y.%
, Louradour, J.%
, Collobert, R.%
\BCBL{}\ \BBA{} Weston, J.%
%
\unskip\
\newblock
\APACrefYearMonthDay{2009}{}{}.
\newblock
\BBOQ{}\APACrefatitle{Curriculum learning}{Curriculum learning}.\BBCQ{}
\newblock
\BIn{} \APACrefbtitle{Proceedings of the 26th annual international conference
  on machine learning}{Proceedings of the 26th annual international conference
  on machine learning}\ (\BPGS\ 41--48).
\PrintBackRefs{\CurrentBib}

\bibitem[\protect\citeauthoryear{%
Blaisdell%
\ \BBA{} Cook%
}{%
Blaisdell%
\ \BBA{} Cook%
}{%
{\protect\APACyear{2005}}%
}]{%
blaisdell2005two}%
\APACinsertmetastar{%
blaisdell2005two}%
Blaisdell, A\BPBI P.%
\BCBT{}\ \BBA{} Cook, R\BPBI G.%
%
\unskip\
\newblock
\APACrefYearMonthDay{2005}{}{}.
\newblock
\BBOQ{}\APACrefatitle{Two-itemsame-different concept learning in
  pigeons}{Two-itemsame-different concept learning in pigeons}.\BBCQ{}
\newblock
\APACjournalVolNumPages{Animal Learning \& Behavior}{33}{1}{67--77}.
\PrintBackRefs{\CurrentBib}

\bibitem[\protect\citeauthoryear{%
{Blender Online Community}%
}{%
{Blender Online Community}%
}{%
{\protect\APACyear{2017}}%
}]{%
blender}%
\APACinsertmetastar{%
blender}%
{Blender Online Community}.%
%
\unskip\
\newblock
\APACrefYearMonthDay{2017}{}{}.
\newblock
\BBOQ{}\APACrefatitle{Blender - a 3D modelling and rendering package}{Blender -
  a 3d modelling and rendering package}\BBCQ{}\ [\bibcomputersoftwaremanual].
\newblock
\APACaddressPublisher{Blender Institute, Amsterdam}{}.
\newblock
 \begin{APACrefURL} \url{http://www.blender.org} \end{APACrefURL}
\PrintBackRefs{\CurrentBib}

\bibitem[\protect\citeauthoryear{%
Bohn%
, Hu%
\BCBL{}\ \BBA{} Ling%
}{%
Bohn%
\ \protect\BOthers{.}}{%
{\protect\APACyear{2019}}%
}]{%
bohn2019few}%
\APACinsertmetastar{%
bohn2019few}%
Bohn, T.%
, Hu, Y.%
\BCBL{}\ \BBA{} Ling, C\BPBI X.%
%
\unskip\
\newblock
\APACrefYearMonthDay{2019}{}{}.
\newblock
\BBOQ{}\APACrefatitle{Few-Shot Abstract Visual Reasoning With Spectral
  Features}{Few-shot abstract visual reasoning with spectral features}.\BBCQ{}
\newblock
\APACjournalVolNumPages{arXiv preprint arXiv:1910.01833}{}{}{}.
\PrintBackRefs{\CurrentBib}

\bibitem[\protect\citeauthoryear{%
Bond%
, Kamil%
, Balda%
\BCBL{}\ \protect\BOthers{.}}{%
Bond%
\ \protect\BOthers{.}}{%
{\protect\APACyear{2004}}%
}]{%
bond2004pinyon}%
\APACinsertmetastar{%
bond2004pinyon}%
Bond, A\BPBI B.%
, Kamil, A\BPBI C.%
, Balda, R\BPBI P.%
\BCBL{}\ \BOthersPeriod{.}%
\unskip\
\newblock
\APACrefYearMonthDay{2004}{}{}.
\newblock
\BBOQ{}\APACrefatitle{Pinyon jays use transitive inference to predict social
  dominance}{Pinyon jays use transitive inference to predict social
  dominance}.\BBCQ{}
\newblock
\APACjournalVolNumPages{Nature}{430}{7001}{778--781}.
\PrintBackRefs{\CurrentBib}

\bibitem[\protect\citeauthoryear{%
Bongard%
}{%
Bongard%
}{%
{\protect\APACyear{1970}}%
}]{%
bongard1970}%
\APACinsertmetastar{%
bongard1970}%
Bongard, M\BPBI M.%
%
\unskip\
\newblock
\APACrefYear{1970}.
\newblock
\APACrefbtitle{Pattern Recognition}{Pattern recognition}.
\newblock
\APACaddressPublisher{}{Spartan Books}.
\PrintBackRefs{\CurrentBib}

\bibitem[\protect\citeauthoryear{%
Bromley%
, Guyon%
, LeCun%
, S{\"a}ckinger%
\BCBL{}\ \BBA{} Shah%
}{%
Bromley%
\ \protect\BOthers{.}}{%
{\protect\APACyear{1994}}%
}]{%
bromley1994signature}%
\APACinsertmetastar{%
bromley1994signature}%
Bromley, J.%
, Guyon, I.%
, LeCun, Y.%
, S{\"a}ckinger, E.%
\BCBL{}\ \BBA{} Shah, R.%
%
\unskip\
\newblock
\APACrefYearMonthDay{1994}{}{}.
\newblock
\BBOQ{}\APACrefatitle{Signature verification using a" siamese" time delay
  neural network}{Signature verification using a" siamese" time delay neural
  network}.\BBCQ{}
\newblock
\BIn{} \APACrefbtitle{Advances in neural information processing
  systems}{Advances in neural information processing systems}\ (\BPGS\
  737--744).
\PrintBackRefs{\CurrentBib}

\bibitem[\protect\citeauthoryear{%
Byosiere%
, Feng%
, Chouinard%
, Howell%
\BCBL{}\ \BBA{} Bennett%
}{%
Byosiere%
\ \protect\BOthers{.}}{%
{\protect\APACyear{2017}}%
}]{%
byosiere2017relational}%
\APACinsertmetastar{%
byosiere2017relational}%
Byosiere, S\BHBI E.%
, Feng, L\BPBI C.%
, Chouinard, P\BPBI A.%
, Howell, T\BPBI J.%
\BCBL{}\ \BBA{} Bennett, P\BPBI C.%
%
\unskip\
\newblock
\APACrefYearMonthDay{2017}{}{}.
\newblock
\BBOQ{}\APACrefatitle{Relational concept learning in domestic dogs: Performance
  on a two-choice size discrimination task generalises to novel
  stimuli}{Relational concept learning in domestic dogs: Performance on a
  two-choice size discrimination task generalises to novel stimuli}.\BBCQ{}
\newblock
\APACjournalVolNumPages{Behavioural processes}{145}{}{93--101}.
\PrintBackRefs{\CurrentBib}

\bibitem[\protect\citeauthoryear{%
Cammarata%
\ \protect\BOthers{.}}{%
Cammarata%
\ \protect\BOthers{.}}{%
{\protect\APACyear{2020}}%
}]{%
cammarata2020thread:}%
\APACinsertmetastar{%
cammarata2020thread:}%
Cammarata, N.%
, Carter, S.%
, Goh, G.%
, Olah, C.%
, Petrov, M.%
\BCBL{}\ \BBA{} Schubert, L.%
%
\unskip\
\newblock
\APACrefYearMonthDay{2020}{}{}.
\newblock
\BBOQ{}\APACrefatitle{Thread: Circuits}{Thread: Circuits}.\BBCQ{}
\newblock
\APACjournalVolNumPages{Distill}{}{}{}.
\newblock
\APACrefnote{https://distill.pub/2020/circuits}
\PrintBackRefs{\CurrentBib}

\bibitem[\protect\citeauthoryear{%
Cho%
\ \protect\BOthers{.}}{%
Cho%
\ \protect\BOthers{.}}{%
{\protect\APACyear{2014}}%
}]{%
cho2014learning}%
\APACinsertmetastar{%
cho2014learning}%
Cho, K.%
, Van~Merri{\"e}nboer, B.%
, Gulcehre, C.%
, Bahdanau, D.%
, Bougares, F.%
, Schwenk, H.%
\BCBL{}\ \BOthersPeriod{.}%
\unskip\
\newblock
\APACrefYearMonthDay{2014}{}{}.
\newblock
\BBOQ{}\APACrefatitle{Learning phrase representations using RNN encoder-decoder
  for statistical machine translation}{Learning phrase representations using
  rnn encoder-decoder for statistical machine translation}.\BBCQ{}
\newblock
\APACjournalVolNumPages{arXiv preprint arXiv:1406.1078}{}{}{}.
\PrintBackRefs{\CurrentBib}

\bibitem[\protect\citeauthoryear{%
Deng%
\ \protect\BOthers{.}}{%
Deng%
\ \protect\BOthers{.}}{%
{\protect\APACyear{2009}}%
}]{%
deng2009imagenet}%
\APACinsertmetastar{%
deng2009imagenet}%
Deng, J.%
, Dong, W.%
, Socher, R.%
, Li, L\BHBI J.%
, Li, K.%
\BCBL{}\ \BBA{} Fei-Fei, L.%
%
\unskip\
\newblock
\APACrefYearMonthDay{2009}{}{}.
\newblock
\BBOQ{}\APACrefatitle{Imagenet: A large-scale hierarchical image
  database}{Imagenet: A large-scale hierarchical image database}.\BBCQ{}
\newblock
\BIn{} \APACrefbtitle{2009 IEEE conference on computer vision and pattern
  recognition}{2009 ieee conference on computer vision and pattern
  recognition}\ (\BPGS\ 248--255).
\PrintBackRefs{\CurrentBib}

\bibitem[\protect\citeauthoryear{%
Depeweg%
, Rothkopf%
\BCBL{}\ \BBA{} J{\"a}kel%
}{%
Depeweg%
\ \protect\BOthers{.}}{%
{\protect\APACyear{2018}}%
}]{%
depeweg2018solving}%
\APACinsertmetastar{%
depeweg2018solving}%
Depeweg, S.%
, Rothkopf, C\BPBI A.%
\BCBL{}\ \BBA{} J{\"a}kel, F.%
%
\unskip\
\newblock
\APACrefYearMonthDay{2018}{}{}.
\newblock
\BBOQ{}\APACrefatitle{Solving Bongard Problems with a Visual Language and
  Pragmatic Reasoning}{Solving bongard problems with a visual language and
  pragmatic reasoning}.\BBCQ{}
\newblock
\APACjournalVolNumPages{arXiv preprint arXiv:1804.04452}{}{}{}.
\PrintBackRefs{\CurrentBib}

\bibitem[\protect\citeauthoryear{%
Devlin%
, Chang%
, Lee%
\BCBL{}\ \BBA{} Toutanova%
}{%
Devlin%
\ \protect\BOthers{.}}{%
{\protect\APACyear{2018}}%
}]{%
bert}%
\APACinsertmetastar{%
bert}%
Devlin, J.%
, Chang, M\BHBI W.%
, Lee, K.%
\BCBL{}\ \BBA{} Toutanova, K.%
%
\unskip\
\newblock
\APACrefYearMonthDay{2018}{}{}.
\newblock
\BBOQ{}\APACrefatitle{Bert: Pre-training of deep bidirectional transformers for
  language understanding}{Bert: Pre-training of deep bidirectional transformers
  for language understanding}.\BBCQ{}
\newblock
\APACjournalVolNumPages{arXiv preprint arXiv:1810.04805}{}{}{}.
\PrintBackRefs{\CurrentBib}

\bibitem[\protect\citeauthoryear{%
Donahue%
\ \protect\BOthers{.}}{%
Donahue%
\ \protect\BOthers{.}}{%
{\protect\APACyear{2015}}%
}]{%
donahue2015long}%
\APACinsertmetastar{%
donahue2015long}%
Donahue, J.%
, Anne~Hendricks, L.%
, Guadarrama, S.%
, Rohrbach, M.%
, Venugopalan, S.%
, Saenko, K.%
\BCBL{}\ \BOthersPeriod{.}%
\unskip\
\newblock
\APACrefYearMonthDay{2015}{}{}.
\newblock
\BBOQ{}\APACrefatitle{Long-term recurrent convolutional networks for visual
  recognition and description}{Long-term recurrent convolutional networks for
  visual recognition and description}.\BBCQ{}
\newblock
\BIn{} \APACrefbtitle{Proceedings of the IEEE conference on computer vision and
  pattern recognition}{Proceedings of the ieee conference on computer vision
  and pattern recognition}\ (\BPGS\ 2625--2634).
\PrintBackRefs{\CurrentBib}

\bibitem[\protect\citeauthoryear{%
Dosovitskiy%
\ \protect\BOthers{.}}{%
Dosovitskiy%
\ \protect\BOthers{.}}{%
{\protect\APACyear{2020}}%
}]{%
visiontransformer}%
\APACinsertmetastar{%
visiontransformer}%
Dosovitskiy, A.%
, Beyer, L.%
, Kolesnikov, A.%
, Weissenborn, D.%
, Zhai, X.%
, Unterthiner, T.%
\BCBL{}\ \BOthersPeriod{.}%
\unskip\
\newblock
\APACrefYearMonthDay{2020}{}{}.
\newblock
\BBOQ{}\APACrefatitle{An image is worth 16x16 words: Transformers for image
  recognition at scale}{An image is worth 16x16 words: Transformers for image
  recognition at scale}.\BBCQ{}
\newblock
\APACjournalVolNumPages{arXiv preprint arXiv:2010.11929}{}{}{}.
\PrintBackRefs{\CurrentBib}

\bibitem[\protect\citeauthoryear{%
Fang%
\ \protect\BOthers{.}}{%
Fang%
\ \protect\BOthers{.}}{%
{\protect\APACyear{2015}}%
}]{%
fang2015captions}%
\APACinsertmetastar{%
fang2015captions}%
Fang, H.%
, Gupta, S.%
, Iandola, F.%
, Srivastava, R\BPBI K.%
, Deng, L.%
, Doll{\'a}r, P.%
\BCBL{}\ \BOthersPeriod{.}%
\unskip\
\newblock
\APACrefYearMonthDay{2015}{}{}.
\newblock
\BBOQ{}\APACrefatitle{From captions to visual concepts and back}{From captions
  to visual concepts and back}.\BBCQ{}
\newblock
\BIn{} \APACrefbtitle{Proceedings of the IEEE conference on computer vision and
  pattern recognition}{Proceedings of the ieee conference on computer vision
  and pattern recognition}\ (\BPGS\ 1473--1482).
\PrintBackRefs{\CurrentBib}

\bibitem[\protect\citeauthoryear{%
Fleuret%
\ \protect\BOthers{.}}{%
Fleuret%
\ \protect\BOthers{.}}{%
{\protect\APACyear{2011}}%
}]{%
fleuret2011comparing}%
\APACinsertmetastar{%
fleuret2011comparing}%
Fleuret, F.%
, Li, T.%
, Dubout, C.%
, Wampler, E\BPBI K.%
, Yantis, S.%
\BCBL{}\ \BBA{} Geman, D.%
%
\unskip\
\newblock
\APACrefYearMonthDay{2011}{}{}.
\newblock
\BBOQ{}\APACrefatitle{Comparing machines and humans on a visual categorization
  test}{Comparing machines and humans on a visual categorization test}.\BBCQ{}
\newblock
\APACjournalVolNumPages{Proceedings of the National Academy of
  Sciences}{108}{43}{17621--17625}.
\PrintBackRefs{\CurrentBib}

\bibitem[\protect\citeauthoryear{%
Freund%
\ \BBA{} Schapire%
}{%
Freund%
\ \BBA{} Schapire%
}{%
{\protect\APACyear{1997}}%
}]{%
freund1997decision}%
\APACinsertmetastar{%
freund1997decision}%
Freund, Y.%
\BCBT{}\ \BBA{} Schapire, R\BPBI E.%
%
\unskip\
\newblock
\APACrefYearMonthDay{1997}{}{}.
\newblock
\BBOQ{}\APACrefatitle{A decision-theoretic generalization of on-line learning
  and an application to boosting}{A decision-theoretic generalization of
  on-line learning and an application to boosting}.\BBCQ{}
\newblock
\APACjournalVolNumPages{Journal of computer and system
  sciences}{55}{1}{119--139}.
\PrintBackRefs{\CurrentBib}

\bibitem[\protect\citeauthoryear{%
Funke%
\ \protect\BOthers{.}}{%
Funke%
\ \protect\BOthers{.}}{%
{\protect\APACyear{2020}}%
}]{%
funke2020notorious}%
\APACinsertmetastar{%
funke2020notorious}%
Funke, C\BPBI M.%
, Borowski, J.%
, Stosio, K.%
, Brendel, W.%
, Wallis, T\BPBI S.%
\BCBL{}\ \BBA{} Bethge, M.%
%
\unskip\
\newblock
\APACrefYearMonthDay{2020}{}{}.
\newblock
\BBOQ{}\APACrefatitle{The Notorious Difficulty of Comparing Human and Machine
  Perception}{The notorious difficulty of comparing human and machine
  perception}.\BBCQ{}
\newblock
\APACjournalVolNumPages{arXiv preprint arXiv:2004.09406}{}{}{}.
\PrintBackRefs{\CurrentBib}

\bibitem[\protect\citeauthoryear{%
Giurfa%
, Zhang%
, Jenett%
, Menzel%
\BCBL{}\ \BBA{} Srinivasan%
}{%
Giurfa%
\ \protect\BOthers{.}}{%
{\protect\APACyear{2001}}%
}]{%
giurfa2001concepts}%
\APACinsertmetastar{%
giurfa2001concepts}%
Giurfa, M.%
, Zhang, S.%
, Jenett, A.%
, Menzel, R.%
\BCBL{}\ \BBA{} Srinivasan, M\BPBI V.%
%
\unskip\
\newblock
\APACrefYearMonthDay{2001}{}{}.
\newblock
\BBOQ{}\APACrefatitle{The concepts of ‘sameness’ and ‘difference’in an
  insect}{The concepts of ‘sameness’ and ‘difference’in an
  insect}.\BBCQ{}
\newblock
\APACjournalVolNumPages{Nature}{410}{6831}{930}.
\PrintBackRefs{\CurrentBib}

\bibitem[\protect\citeauthoryear{%
Goh%
\ \protect\BOthers{.}}{%
Goh%
\ \protect\BOthers{.}}{%
{\protect\APACyear{2021}}%
}]{%
distill_multimodal}%
\APACinsertmetastar{%
distill_multimodal}%
Goh, G.%
, †, N\BPBI C.%
, †, C\BPBI V.%
, Carter, S.%
, Petrov, M.%
, Schubert, L.%
\BCBL{}\ \BOthersPeriod{.}%
\unskip\
\newblock
\APACrefYearMonthDay{2021}{}{}.
\newblock
\BBOQ{}\APACrefatitle{Multimodal Neurons in Artificial Neural
  Networks}{Multimodal neurons in artificial neural networks}.\BBCQ{}
\newblock
\APACjournalVolNumPages{Distill}{}{}{}.
\newblock
https://distill.pub/2021/multimodal-neurons, doi:10.23915/distill.00030
\newblock
[\href{https://distill.pub/2021/multimodal-neurons}{Article}]
\PrintBackRefs{\CurrentBib}

\bibitem[\protect\citeauthoryear{%
Grosenick%
, Clement%
\BCBL{}\ \BBA{} Fernald%
}{%
Grosenick%
\ \protect\BOthers{.}}{%
{\protect\APACyear{2007}}%
}]{%
grosenick2007fish}%
\APACinsertmetastar{%
grosenick2007fish}%
Grosenick, L.%
, Clement, T\BPBI S.%
\BCBL{}\ \BBA{} Fernald, R\BPBI D.%
%
\unskip\
\newblock
\APACrefYearMonthDay{2007}{}{}.
\newblock
\BBOQ{}\APACrefatitle{Fish can infer social rank by observation alone}{Fish can
  infer social rank by observation alone}.\BBCQ{}
\newblock
\APACjournalVolNumPages{Nature}{445}{7126}{429--432}.
\PrintBackRefs{\CurrentBib}

\bibitem[\protect\citeauthoryear{%
He%
, Zhang%
, Ren%
\BCBL{}\ \BBA{} Sun%
}{%
He%
\ \protect\BOthers{.}}{%
{\protect\APACyear{2015}}%
}]{%
he2015delving}%
\APACinsertmetastar{%
he2015delving}%
He, K.%
, Zhang, X.%
, Ren, S.%
\BCBL{}\ \BBA{} Sun, J.%
%
\unskip\
\newblock
\APACrefYearMonthDay{2015}{}{}.
\newblock
\BBOQ{}\APACrefatitle{Delving deep into rectifiers: Surpassing human-level
  performance on imagenet classification}{Delving deep into rectifiers:
  Surpassing human-level performance on imagenet classification}.\BBCQ{}
\newblock
\BIn{} \APACrefbtitle{Proceedings of the IEEE international conference on
  computer vision}{Proceedings of the ieee international conference on computer
  vision}\ (\BPGS\ 1026--1034).
\PrintBackRefs{\CurrentBib}

\bibitem[\protect\citeauthoryear{%
He%
, Zhang%
, Ren%
\BCBL{}\ \BBA{} Sun%
}{%
He%
\ \protect\BOthers{.}}{%
{\protect\APACyear{2016}}%
}]{%
he2016deep}%
\APACinsertmetastar{%
he2016deep}%
He, K.%
, Zhang, X.%
, Ren, S.%
\BCBL{}\ \BBA{} Sun, J.%
%
\unskip\
\newblock
\APACrefYearMonthDay{2016}{}{}.
\newblock
\BBOQ{}\APACrefatitle{Deep residual learning for image recognition}{Deep
  residual learning for image recognition}.\BBCQ{}
\newblock
\BIn{} \APACrefbtitle{Proceedings of the IEEE conference on computer vision and
  pattern recognition}{Proceedings of the ieee conference on computer vision
  and pattern recognition}\ (\BPGS\ 770--778).
\PrintBackRefs{\CurrentBib}

\bibitem[\protect\citeauthoryear{%
Herrnstein%
\ \BBA{} Loveland%
}{%
Herrnstein%
\ \BBA{} Loveland%
}{%
{\protect\APACyear{1964}}%
}]{%
herrnstein1964complex}%
\APACinsertmetastar{%
herrnstein1964complex}%
Herrnstein, R\BPBI J.%
\BCBT{}\ \BBA{} Loveland, D\BPBI H.%
%
\unskip\
\newblock
\APACrefYearMonthDay{1964}{}{}.
\newblock
\BBOQ{}\APACrefatitle{Complex visual concept in the pigeon}{Complex visual
  concept in the pigeon}.\BBCQ{}
\newblock
\APACjournalVolNumPages{Science}{146}{3643}{549--551}.
\PrintBackRefs{\CurrentBib}

\bibitem[\protect\citeauthoryear{%
Hochreiter%
\ \BBA{} Schmidhuber%
}{%
Hochreiter%
\ \BBA{} Schmidhuber%
}{%
{\protect\APACyear{1997}}%
}]{%
hochreiter1997long}%
\APACinsertmetastar{%
hochreiter1997long}%
Hochreiter, S.%
\BCBT{}\ \BBA{} Schmidhuber, J.%
%
\unskip\
\newblock
\APACrefYearMonthDay{1997}{}{}.
\newblock
\BBOQ{}\APACrefatitle{Long short-term memory}{Long short-term memory}.\BBCQ{}
\newblock
\APACjournalVolNumPages{Neural computation}{9}{8}{1735--1780}.
\PrintBackRefs{\CurrentBib}

\bibitem[\protect\citeauthoryear{%
Hofstadter%
}{%
Hofstadter%
}{%
{\protect\APACyear{1979}}%
}]{%
douglas1979godel}%
\APACinsertmetastar{%
douglas1979godel}%
Hofstadter, D\BPBI R.%
%
\unskip\
\newblock
\APACrefYearMonthDay{1979}{}{}.
\newblock
\APACrefbtitle{G{\"o}del, escher, bach: An eternal golden braid.}{G{\"o}del,
  escher, bach: An eternal golden braid.}
\newblock
\APACaddressPublisher{}{Basic Books, New York}.
\PrintBackRefs{\CurrentBib}

\bibitem[\protect\citeauthoryear{%
Hogue%
, Beaugrand%
\BCBL{}\ \BBA{} Lagu{\"e}%
}{%
Hogue%
\ \protect\BOthers{.}}{%
{\protect\APACyear{1996}}%
}]{%
hogue1996coherent}%
\APACinsertmetastar{%
hogue1996coherent}%
Hogue, M\BHBI E.%
, Beaugrand, J\BPBI P.%
\BCBL{}\ \BBA{} Lagu{\"e}, P\BPBI C.%
%
\unskip\
\newblock
\APACrefYearMonthDay{1996}{}{}.
\newblock
\BBOQ{}\APACrefatitle{Coherent use of information by hens observing their
  former dominant defeating or being defeated by a stranger}{Coherent use of
  information by hens observing their former dominant defeating or being
  defeated by a stranger}.\BBCQ{}
\newblock
\APACjournalVolNumPages{Behavioural Processes}{38}{3}{241--252}.
\PrintBackRefs{\CurrentBib}

\bibitem[\protect\citeauthoryear{%
Hoshen%
\ \BBA{} Werman%
}{%
Hoshen%
\ \BBA{} Werman%
}{%
{\protect\APACyear{2017}}%
}]{%
hoshen2017iq}%
\APACinsertmetastar{%
hoshen2017iq}%
Hoshen, D.%
\BCBT{}\ \BBA{} Werman, M.%
%
\unskip\
\newblock
\APACrefYearMonthDay{2017}{}{}.
\newblock
\BBOQ{}\APACrefatitle{IQ of Neural Networks}{Iq of neural networks}.\BBCQ{}
\newblock
\APACjournalVolNumPages{arXiv preprint arXiv:1710.01692}{}{}{}.
\PrintBackRefs{\CurrentBib}

\bibitem[\protect\citeauthoryear{%
Jaegle%
\ \protect\BOthers{.}}{%
Jaegle%
\ \protect\BOthers{.}}{%
{\protect\APACyear{2021}}%
}]{%
perceiver}%
\APACinsertmetastar{%
perceiver}%
Jaegle, A.%
, Gimeno, F.%
, Brock, A.%
, Zisserman, A.%
, Vinyals, O.%
\BCBL{}\ \BBA{} Carreira, J.%
%
\unskip\
\newblock
\APACrefYearMonthDay{2021}{}{}.
\newblock
\BBOQ{}\APACrefatitle{Perceiver: General perception with iterative
  attention}{Perceiver: General perception with iterative attention}.\BBCQ{}
\newblock
\APACjournalVolNumPages{arXiv preprint arXiv:2103.03206}{}{}{}.
\PrintBackRefs{\CurrentBib}

\bibitem[\protect\citeauthoryear{%
Johnson%
\ \protect\BOthers{.}}{%
Johnson%
\ \protect\BOthers{.}}{%
{\protect\APACyear{2017}}%
}]{%
johnson2017clevr}%
\APACinsertmetastar{%
johnson2017clevr}%
Johnson, J.%
, Hariharan, B.%
, Maaten, L. van~der%
, Fei-Fei, L.%
, Lawrence~Zitnick, C.%
\BCBL{}\ \BBA{} Girshick, R.%
%
\unskip\
\newblock
\APACrefYearMonthDay{2017}{}{}.
\newblock
\BBOQ{}\APACrefatitle{Clevr: A diagnostic dataset for compositional language
  and elementary visual reasoning}{Clevr: A diagnostic dataset for
  compositional language and elementary visual reasoning}.\BBCQ{}
\newblock
\BIn{} \APACrefbtitle{Proceedings of the IEEE Conference on Computer Vision and
  Pattern Recognition}{Proceedings of the ieee conference on computer vision
  and pattern recognition}\ (\BPGS\ 2901--2910).
\PrintBackRefs{\CurrentBib}

\bibitem[\protect\citeauthoryear{%
Karpathy%
}{%
Karpathy%
}{%
{\protect\APACyear{2014}}%
}]{%
imagenet_human_subject}%
\APACinsertmetastar{%
imagenet_human_subject}%
Karpathy, A.%
%
\unskip\
\newblock
\APACrefYearMonthDay{2014}{Sep}{}.
\newblock

\PrintBackRefs{\CurrentBib}

\bibitem[\protect\citeauthoryear{%
Katz%
\ \BBA{} Wright%
}{%
Katz%
\ \BBA{} Wright%
}{%
{\protect\APACyear{2006}}%
}]{%
katz2006same}%
\APACinsertmetastar{%
katz2006same}%
Katz, J\BPBI S.%
\BCBT{}\ \BBA{} Wright, A\BPBI A.%
%
\unskip\
\newblock
\APACrefYearMonthDay{2006}{}{}.
\newblock
\BBOQ{}\APACrefatitle{Same/different abstract-concept learning by
  pigeons.}{Same/different abstract-concept learning by pigeons.}\BBCQ{}
\newblock
\APACjournalVolNumPages{Journal of Experimental Psychology: Animal Behavior
  Processes}{32}{1}{80}.
\PrintBackRefs{\CurrentBib}

\bibitem[\protect\citeauthoryear{%
E.~Kim%
, Hannan%
\BCBL{}\ \BBA{} Kenyon%
}{%
E.~Kim%
\ \protect\BOthers{.}}{%
{\protect\APACyear{2018}}%
}]{%
halle_berry_neurons}%
\APACinsertmetastar{%
halle_berry_neurons}%
Kim, E.%
, Hannan, D.%
\BCBL{}\ \BBA{} Kenyon, G.%
%
\unskip\
\newblock
\APACrefYearMonthDay{2018}{June}{}.
\newblock
\BBOQ{}\APACrefatitle{Deep Sparse Coding for Invariant Multimodal Halle Berry
  Neurons}{Deep sparse coding for invariant multimodal halle berry
  neurons}.\BBCQ{}
\newblock
\BIn{} \APACrefbtitle{Proceedings of the IEEE Conference on Computer Vision and
  Pattern Recognition (CVPR).}{Proceedings of the ieee conference on computer
  vision and pattern recognition (cvpr).}
\PrintBackRefs{\CurrentBib}

\bibitem[\protect\citeauthoryear{%
J.~Kim%
, Ricci%
\BCBL{}\ \BBA{} Serre%
}{%
J.~Kim%
\ \protect\BOthers{.}}{%
{\protect\APACyear{2018}}%
}]{%
kim2018not}%
\APACinsertmetastar{%
kim2018not}%
Kim, J.%
, Ricci, M.%
\BCBL{}\ \BBA{} Serre, T.%
%
\unskip\
\newblock
\APACrefYearMonthDay{2018}{}{}.
\newblock
\BBOQ{}\APACrefatitle{Not-So-CLEVR: learning same--different relations strains
  feedforward neural networks}{Not-so-clevr: learning same--different relations
  strains feedforward neural networks}.\BBCQ{}
\newblock
\APACjournalVolNumPages{Interface Focus}{8}{4}{20180011}.
\PrintBackRefs{\CurrentBib}

\bibitem[\protect\citeauthoryear{%
Krizhevsky%
, Sutskever%
\BCBL{}\ \BBA{} Hinton%
}{%
Krizhevsky%
\ \protect\BOthers{.}}{%
{\protect\APACyear{2012}}%
}]{%
krizhevsky2012imagenet}%
\APACinsertmetastar{%
krizhevsky2012imagenet}%
Krizhevsky, A.%
, Sutskever, I.%
\BCBL{}\ \BBA{} Hinton, G\BPBI E.%
%
\unskip\
\newblock
\APACrefYearMonthDay{2012}{}{}.
\newblock
\BBOQ{}\APACrefatitle{Imagenet classification with deep convolutional neural
  networks}{Imagenet classification with deep convolutional neural
  networks}.\BBCQ{}
\newblock
\BIn{} \APACrefbtitle{Advances in neural information processing
  systems}{Advances in neural information processing systems}\ (\BPGS\
  1097--1105).
\PrintBackRefs{\CurrentBib}

\bibitem[\protect\citeauthoryear{%
Kubilius%
\ \protect\BOthers{.}}{%
Kubilius%
\ \protect\BOthers{.}}{%
{\protect\APACyear{2018}}%
}]{%
kubilius2018cornet}%
\APACinsertmetastar{%
kubilius2018cornet}%
Kubilius, J.%
, Schrimpf, M.%
, Nayebi, A.%
, Bear, D.%
, Yamins, D\BPBI L.%
\BCBL{}\ \BBA{} DiCarlo, J\BPBI J.%
%
\unskip\
\newblock
\APACrefYearMonthDay{2018}{}{}.
\newblock
\BBOQ{}\APACrefatitle{CORnet: modeling the neural mechanisms of core object
  recognition}{Cornet: modeling the neural mechanisms of core object
  recognition}.\BBCQ{}
\newblock
\APACjournalVolNumPages{BioRxiv}{}{}{408385}.
\PrintBackRefs{\CurrentBib}

\bibitem[\protect\citeauthoryear{%
LeCun%
, Bengio%
\BCBL{}\ \BBA{} Hinton%
}{%
LeCun%
\ \protect\BOthers{.}}{%
{\protect\APACyear{2015}}%
}]{%
lecun2015deep}%
\APACinsertmetastar{%
lecun2015deep}%
LeCun, Y.%
, Bengio, Y.%
\BCBL{}\ \BBA{} Hinton, G.%
%
\unskip\
\newblock
\APACrefYearMonthDay{2015}{}{}.
\newblock
\BBOQ{}\APACrefatitle{Deep learning}{Deep learning}.\BBCQ{}
\newblock
\APACjournalVolNumPages{nature}{521}{7553}{436}.
\PrintBackRefs{\CurrentBib}

\bibitem[\protect\citeauthoryear{%
LeCun%
\ \protect\BOthers{.}}{%
LeCun%
\ \protect\BOthers{.}}{%
{\protect\APACyear{1989}}%
}]{%
lecun1989backpropagation}%
\APACinsertmetastar{%
lecun1989backpropagation}%
LeCun, Y.%
, Boser, B.%
, Denker, J\BPBI S.%
, Henderson, D.%
, Howard, R\BPBI E.%
, Hubbard, W.%
\BCBL{}\ \BOthersPeriod{.}%
\unskip\
\newblock
\APACrefYearMonthDay{1989}{}{}.
\newblock
\BBOQ{}\APACrefatitle{Backpropagation applied to handwritten zip code
  recognition}{Backpropagation applied to handwritten zip code
  recognition}.\BBCQ{}
\newblock
\APACjournalVolNumPages{Neural computation}{1}{4}{541--551}.
\PrintBackRefs{\CurrentBib}

\bibitem[\protect\citeauthoryear{%
M{\aa}rtensson%
\ \protect\BOthers{.}}{%
M{\aa}rtensson%
\ \protect\BOthers{.}}{%
{\protect\APACyear{2020}}%
}]{%
mri-out-of-distribution}%
\APACinsertmetastar{%
mri-out-of-distribution}%
M{\aa}rtensson, G.%
, Ferreira, D.%
, Granberg, T.%
, Cavallin, L.%
, Oppedal, K.%
, Padovani, A.%
\BCBL{}\ \BOthersPeriod{.}%
\unskip\
\newblock
\APACrefYearMonthDay{2020}{}{}.
\newblock
\BBOQ{}\APACrefatitle{The reliability of a deep learning model in clinical
  out-of-distribution MRI data: a multicohort study}{The reliability of a deep
  learning model in clinical out-of-distribution mri data: a multicohort
  study}.\BBCQ{}
\newblock
\APACjournalVolNumPages{Medical Image Analysis}{66}{}{101714}.
\PrintBackRefs{\CurrentBib}

\bibitem[\protect\citeauthoryear{%
Martinho%
\ \BBA{} Kacelnik%
}{%
Martinho%
\ \BBA{} Kacelnik%
}{%
{\protect\APACyear{2016}}%
}]{%
martinho2016ducklings}%
\APACinsertmetastar{%
martinho2016ducklings}%
Martinho, A.%
\BCBT{}\ \BBA{} Kacelnik, A.%
%
\unskip\
\newblock
\APACrefYearMonthDay{2016}{}{}.
\newblock
\BBOQ{}\APACrefatitle{Ducklings imprint on the relational concept of “same or
  different”}{Ducklings imprint on the relational concept of “same or
  different”}.\BBCQ{}
\newblock
\APACjournalVolNumPages{Science}{353}{6296}{286--288}.
\PrintBackRefs{\CurrentBib}

\bibitem[\protect\citeauthoryear{%
Mercado~III%
, Killebrew%
, Pack%
, M{\'a}cha%
\BCBL{}\ \BBA{} Herman%
}{%
Mercado~III%
\ \protect\BOthers{.}}{%
{\protect\APACyear{2000}}%
}]{%
mercado2000generalization}%
\APACinsertmetastar{%
mercado2000generalization}%
Mercado~III, E.%
, Killebrew, D\BPBI A.%
, Pack, A\BPBI A.%
, M{\'a}cha, I\BPBI V.%
\BCBL{}\ \BBA{} Herman, L\BPBI M.%
%
\unskip\
\newblock
\APACrefYearMonthDay{2000}{}{}.
\newblock
\BBOQ{}\APACrefatitle{Generalization of ‘same--different’classification
  abilities in bottlenosed dolphins}{Generalization of
  ‘same--different’classification abilities in bottlenosed
  dolphins}.\BBCQ{}
\newblock
\APACjournalVolNumPages{Behavioural Processes}{50}{2-3}{79--94}.
\PrintBackRefs{\CurrentBib}

\bibitem[\protect\citeauthoryear{%
Messina%
, Amato%
, Carrara%
, Falchi%
\BCBL{}\ \BBA{} Gennaro%
}{%
Messina%
\ \protect\BOthers{.}}{%
{\protect\APACyear{2019}}%
}]{%
messina2019testing}%
\APACinsertmetastar{%
messina2019testing}%
Messina, N.%
, Amato, G.%
, Carrara, F.%
, Falchi, F.%
\BCBL{}\ \BBA{} Gennaro, C.%
%
\unskip\
\newblock
\APACrefYearMonthDay{2019}{}{}.
\newblock
\BBOQ{}\APACrefatitle{Testing Deep Neural Networks on the Same-Different
  Task}{Testing deep neural networks on the same-different task}.\BBCQ{}
\newblock
\BIn{} \APACrefbtitle{2019 International Conference on Content-Based Multimedia
  Indexing (CBMI)}{2019 international conference on content-based multimedia
  indexing (cbmi)}\ (\BPGS\ 1--6).
\PrintBackRefs{\CurrentBib}

\bibitem[\protect\citeauthoryear{%
Mondrag{\'o}n%
, Alonso%
\BCBL{}\ \BBA{} Kokkola%
}{%
Mondrag{\'o}n%
\ \protect\BOthers{.}}{%
{\protect\APACyear{2017}}%
}]{%
mondragon2017associative}%
\APACinsertmetastar{%
mondragon2017associative}%
Mondrag{\'o}n, E.%
, Alonso, E.%
\BCBL{}\ \BBA{} Kokkola, N.%
%
\unskip\
\newblock
\APACrefYearMonthDay{2017}{}{}.
\newblock
\BBOQ{}\APACrefatitle{Associative learning should go deep}{Associative learning
  should go deep}.\BBCQ{}
\newblock
\APACjournalVolNumPages{Trends in cognitive sciences}{21}{11}{822--825}.
\PrintBackRefs{\CurrentBib}

\bibitem[\protect\citeauthoryear{%
Murphy%
}{%
Murphy%
}{%
{\protect\APACyear{2004}}%
}]{%
bigbook}%
\APACinsertmetastar{%
bigbook}%
Murphy, G.%
%
\unskip\
\newblock
\APACrefYear{2004}.
\newblock
\APACrefbtitle{The big book of concepts}{The big book of concepts}.
\newblock
\APACaddressPublisher{}{MIT press}.
\PrintBackRefs{\CurrentBib}

\bibitem[\protect\citeauthoryear{%
Nie%
\ \protect\BOthers{.}}{%
Nie%
\ \protect\BOthers{.}}{%
{\protect\APACyear{2020}}%
}]{%
nie2020bongard}%
\APACinsertmetastar{%
nie2020bongard}%
Nie, W.%
, Yu, Z.%
, Mao, L.%
, Patel, A\BPBI B.%
, Zhu, Y.%
\BCBL{}\ \BBA{} Anandkumar, A.%
%
\unskip\
\newblock
\APACrefYearMonthDay{2020}{}{}.
\newblock
\BBOQ{}\APACrefatitle{BONGARD-LOGO: A New Benchmark for Human-Level Concept
  Learning and Reasoning}{Bongard-logo: A new benchmark for human-level concept
  learning and reasoning}.\BBCQ{}
\newblock
\APACjournalVolNumPages{arXiv preprint arXiv:2010.00763}{}{}{}.
\PrintBackRefs{\CurrentBib}

\bibitem[\protect\citeauthoryear{%
Novak%
, Bahri%
, Abolafia%
, Pennington%
\BCBL{}\ \BBA{} Sohl-Dickstein%
}{%
Novak%
\ \protect\BOthers{.}}{%
{\protect\APACyear{2018}}%
}]{%
sensitivity}%
\APACinsertmetastar{%
sensitivity}%
Novak, R.%
, Bahri, Y.%
, Abolafia, D\BPBI A.%
, Pennington, J.%
\BCBL{}\ \BBA{} Sohl-Dickstein, J.%
%
\unskip\
\newblock
\APACrefYearMonthDay{2018}{}{}.
\newblock
\BBOQ{}\APACrefatitle{Sensitivity and generalization in neural networks: an
  empirical study}{Sensitivity and generalization in neural networks: an
  empirical study}.\BBCQ{}
\newblock
\APACjournalVolNumPages{arXiv preprint arXiv:1802.08760}{}{}{}.
\PrintBackRefs{\CurrentBib}

\bibitem[\protect\citeauthoryear{%
Oden%
, Thompson%
\BCBL{}\ \BBA{} Premack%
}{%
Oden%
\ \protect\BOthers{.}}{%
{\protect\APACyear{1990}}%
}]{%
oden1990infant}%
\APACinsertmetastar{%
oden1990infant}%
Oden, D\BPBI L.%
, Thompson, R\BPBI K.%
\BCBL{}\ \BBA{} Premack, D.%
%
\unskip\
\newblock
\APACrefYearMonthDay{1990}{}{}.
\newblock
\BBOQ{}\APACrefatitle{Infant chimpanzees spontaneously perceive both concrete
  and abstract same/different relations}{Infant chimpanzees spontaneously
  perceive both concrete and abstract same/different relations}.\BBCQ{}
\newblock
\APACjournalVolNumPages{Child Development}{61}{3}{621--631}.
\PrintBackRefs{\CurrentBib}

\bibitem[\protect\citeauthoryear{%
Peer%
, Stabinger%
\BCBL{}\ \BBA{} Rodr{\'{\i}}guez{-}S{\'{a}}nchez%
}{%
Peer%
\ \protect\BOthers{.}}{%
{\protect\APACyear{2021}}%
}]{%
cb-autotune}%
\APACinsertmetastar{%
cb-autotune}%
Peer, D.%
, Stabinger, S.%
\BCBL{}\ \BBA{} Rodr{\'{\i}}guez{-}S{\'{a}}nchez, A\BPBI J.%
%
\unskip\
\newblock
\APACrefYearMonthDay{2021}{}{}.
\newblock
\BBOQ{}\APACrefatitle{Auto-tuning of Deep Neural Networks by Conflicting Layer
  Removal}{Auto-tuning of deep neural networks by conflicting layer
  removal}.\BBCQ{}
\newblock
\APACjournalVolNumPages{arXiv preprint arXiv:2103.04331}{}{}{}.
\PrintBackRefs{\CurrentBib}

\bibitem[\protect\citeauthoryear{%
Pepperberg%
}{%
Pepperberg%
}{%
{\protect\APACyear{1987}}%
}]{%
pepperberg1987acquisition}%
\APACinsertmetastar{%
pepperberg1987acquisition}%
Pepperberg, I\BPBI M.%
%
\unskip\
\newblock
\APACrefYearMonthDay{1987}{}{}.
\newblock
\BBOQ{}\APACrefatitle{Acquisition of the same/different concept by an African
  Grey parrot (Psittacus erithacus): Learning with respect to categories of
  color, shape, and material}{Acquisition of the same/different concept by an
  african grey parrot (psittacus erithacus): Learning with respect to
  categories of color, shape, and material}.\BBCQ{}
\newblock
\APACjournalVolNumPages{Animal Learning \& Behavior}{15}{4}{423--432}.
\PrintBackRefs{\CurrentBib}

\bibitem[\protect\citeauthoryear{%
Pham%
, Dai%
, Xie%
, Luong%
\BCBL{}\ \BBA{} Le%
}{%
Pham%
\ \protect\BOthers{.}}{%
{\protect\APACyear{2020}}%
}]{%
pham2020meta}%
\APACinsertmetastar{%
pham2020meta}%
Pham, H.%
, Dai, Z.%
, Xie, Q.%
, Luong, M\BHBI T.%
\BCBL{}\ \BBA{} Le, Q\BPBI V.%
%
\unskip\
\newblock
\APACrefYearMonthDay{2020}{}{}.
\newblock
\BBOQ{}\APACrefatitle{Meta pseudo labels}{Meta pseudo labels}.\BBCQ{}
\newblock
\APACjournalVolNumPages{arXiv preprint arXiv:2003.10580}{}{}{}.
\PrintBackRefs{\CurrentBib}

\bibitem[\protect\citeauthoryear{%
Radford%
\ \protect\BOthers{.}}{%
Radford%
\ \protect\BOthers{.}}{%
{\protect\APACyear{2021}}%
}]{%
clip}%
\APACinsertmetastar{%
clip}%
Radford, A.%
, Kim, J\BPBI W.%
, Hallacy, C.%
, Ramesh, A.%
, Goh, G.%
, Agarwal, S.%
\BCBL{}\ \BOthersPeriod{.}%
\unskip\
\newblock
\APACrefYearMonthDay{2021}{}{}.
\newblock
\BBOQ{}\APACrefatitle{Learning transferable visual models from natural language
  supervision}{Learning transferable visual models from natural language
  supervision}.\BBCQ{}
\newblock
\APACjournalVolNumPages{arXiv preprint arXiv:2103.00020}{}{}{}.
\PrintBackRefs{\CurrentBib}

\bibitem[\protect\citeauthoryear{%
Radford%
, Metz%
\BCBL{}\ \BBA{} Chintala%
}{%
Radford%
\ \protect\BOthers{.}}{%
{\protect\APACyear{2015}}%
}]{%
radford2015unsupervised}%
\APACinsertmetastar{%
radford2015unsupervised}%
Radford, A.%
, Metz, L.%
\BCBL{}\ \BBA{} Chintala, S.%
%
\unskip\
\newblock
\APACrefYearMonthDay{2015}{}{}.
\newblock
\BBOQ{}\APACrefatitle{Unsupervised representation learning with deep
  convolutional generative adversarial networks}{Unsupervised representation
  learning with deep convolutional generative adversarial networks}.\BBCQ{}
\newblock
\APACjournalVolNumPages{arXiv preprint arXiv:1511.06434}{}{}{}.
\PrintBackRefs{\CurrentBib}

\bibitem[\protect\citeauthoryear{%
Raghu%
, Poole%
, Kleinberg%
, Ganguli%
\BCBL{}\ \BBA{} Sohl-Dickstein%
}{%
Raghu%
\ \protect\BOthers{.}}{%
{\protect\APACyear{2017}}%
}]{%
expressivity}%
\APACinsertmetastar{%
expressivity}%
Raghu, M.%
, Poole, B.%
, Kleinberg, J.%
, Ganguli, S.%
\BCBL{}\ \BBA{} Sohl-Dickstein, J.%
%
\unskip\
\newblock
\APACrefYearMonthDay{2017}{}{}.
\newblock
\BBOQ{}\APACrefatitle{On the expressive power of deep neural networks}{On the
  expressive power of deep neural networks}.\BBCQ{}
\newblock
\BIn{} \APACrefbtitle{international conference on machine
  learning}{international conference on machine learning}\ (\BPGS\ 2847--2854).
\PrintBackRefs{\CurrentBib}

\bibitem[\protect\citeauthoryear{%
Raven%
\ \protect\BOthers{.}}{%
Raven%
\ \protect\BOthers{.}}{%
{\protect\APACyear{1938}}%
}]{%
raven1938raven}%
\APACinsertmetastar{%
raven1938raven}%
Raven, J\BPBI C.%
\BCBT{}\ \BOthersPeriod{.}%
\unskip\
\newblock
\APACrefYear{1938}.
\newblock
\APACrefbtitle{Raven's progressive matrices}{Raven's progressive matrices}.
\newblock
\APACaddressPublisher{}{Western Psychological Services Los Angeles, CA}.
\PrintBackRefs{\CurrentBib}

\bibitem[\protect\citeauthoryear{%
Ricci%
, Kim%
\BCBL{}\ \BBA{} Serre%
}{%
Ricci%
\ \protect\BOthers{.}}{%
{\protect\APACyear{2018}}%
{\protect\APACexlab{{\protect\BCnt{1}}}}}]{%
ricci2018not}%
\APACinsertmetastar{%
ricci2018not}%
Ricci, M.%
, Kim, J.%
\BCBL{}\ \BBA{} Serre, T.%
%
\unskip\
\newblock
\APACrefYearMonthDay{2018{\protect\BCnt{1}}}{}{}.
\newblock
\BBOQ{}\APACrefatitle{Not-So-CLEVR: Visual Relations Strain Feedforward Neural
  Networks}{Not-so-clevr: Visual relations strain feedforward neural
  networks}.\BBCQ{}
\newblock
\APACjournalVolNumPages{arXiv preprint arXiv:1802.03390}{}{}{}.
\PrintBackRefs{\CurrentBib}

\bibitem[\protect\citeauthoryear{%
Ricci%
, Kim%
\BCBL{}\ \BBA{} Serre%
}{%
Ricci%
\ \protect\BOthers{.}}{%
{\protect\APACyear{2018}}%
{\protect\APACexlab{{\protect\BCnt{2}}}}}]{%
ricci2018same}%
\APACinsertmetastar{%
ricci2018same}%
Ricci, M.%
, Kim, J.%
\BCBL{}\ \BBA{} Serre, T.%
%
\unskip\
\newblock
\APACrefYearMonthDay{2018{\protect\BCnt{2}}}{}{}.
\newblock
\BBOQ{}\APACrefatitle{Same-different problems strain convolutional neural
  networks}{Same-different problems strain convolutional neural
  networks}.\BBCQ{}
\newblock
\APACjournalVolNumPages{ArXiv180203390 Cs Q-Bio Available at: http://arxiv.
  org/abs/1802.03390 [Accessed May 28, 2018]}{}{}{}.
\PrintBackRefs{\CurrentBib}

\bibitem[\protect\citeauthoryear{%
Rozell%
, Johnson%
, Baraniuk%
\BCBL{}\ \BBA{} Olshausen%
}{%
Rozell%
\ \protect\BOthers{.}}{%
{\protect\APACyear{2007}}%
}]{%
lca}%
\APACinsertmetastar{%
lca}%
Rozell, C.%
, Johnson, D.%
, Baraniuk, R.%
\BCBL{}\ \BBA{} Olshausen, B.%
%
\unskip\
\newblock
\APACrefYearMonthDay{2007}{}{}.
\newblock
\BBOQ{}\APACrefatitle{Locally competitive algorithms for sparse
  approximation}{Locally competitive algorithms for sparse
  approximation}.\BBCQ{}
\newblock
\BIn{} \APACrefbtitle{2007 IEEE International Conference on Image
  Processing}{2007 ieee international conference on image processing}\
  (\BVOL~4, \BPGS\ IV--169).
\PrintBackRefs{\CurrentBib}

\bibitem[\protect\citeauthoryear{%
Russakovsky%
\ \protect\BOthers{.}}{%
Russakovsky%
\ \protect\BOthers{.}}{%
{\protect\APACyear{2015}}%
}]{%
imagenet}%
\APACinsertmetastar{%
imagenet}%
Russakovsky, O.%
, Deng, J.%
, Su, H.%
, Krause, J.%
, Satheesh, S.%
, Ma, S.%
\BCBL{}\ \BOthersPeriod{.}%
\unskip\
\newblock
\APACrefYearMonthDay{2015}{}{}.
\newblock
\BBOQ{}\APACrefatitle{Imagenet large scale visual recognition
  challenge}{Imagenet large scale visual recognition challenge}.\BBCQ{}
\newblock
\APACjournalVolNumPages{International journal of computer
  vision}{115}{3}{211--252}.
\PrintBackRefs{\CurrentBib}

\bibitem[\protect\citeauthoryear{%
Santoro%
\ \protect\BOthers{.}}{%
Santoro%
\ \protect\BOthers{.}}{%
{\protect\APACyear{2017}}%
}]{%
santoro2017simple}%
\APACinsertmetastar{%
santoro2017simple}%
Santoro, A.%
, Raposo, D.%
, Barrett, D\BPBI G.%
, Malinowski, M.%
, Pascanu, R.%
, Battaglia, P.%
\BCBL{}\ \BOthersPeriod{.}%
\unskip\
\newblock
\APACrefYearMonthDay{2017}{}{}.
\newblock
\BBOQ{}\APACrefatitle{A simple neural network module for relational
  reasoning}{A simple neural network module for relational reasoning}.\BBCQ{}
\newblock
\BIn{} \APACrefbtitle{Advances in neural information processing
  systems}{Advances in neural information processing systems}\ (\BPGS\
  4967--4976).
\PrintBackRefs{\CurrentBib}

\bibitem[\protect\citeauthoryear{%
Schaffer%
}{%
Schaffer%
}{%
{\protect\APACyear{2015}}%
}]{%
schaffer2015not}%
\APACinsertmetastar{%
schaffer2015not}%
Schaffer, J.%
%
\unskip\
\newblock
\APACrefYearMonthDay{2015}{}{}.
\newblock
\BBOQ{}\APACrefatitle{What not to multiply without necessity}{What not to
  multiply without necessity}.\BBCQ{}
\newblock
\APACjournalVolNumPages{Australasian Journal of Philosophy}{93}{4}{644--664}.
\PrintBackRefs{\CurrentBib}

\bibitem[\protect\citeauthoryear{%
Schrier%
\ \BBA{} Brady%
}{%
Schrier%
\ \BBA{} Brady%
}{%
{\protect\APACyear{1987}}%
}]{%
schrier1987categorization}%
\APACinsertmetastar{%
schrier1987categorization}%
Schrier, A\BPBI M.%
\BCBT{}\ \BBA{} Brady, P\BPBI M.%
%
\unskip\
\newblock
\APACrefYearMonthDay{1987}{}{}.
\newblock
\BBOQ{}\APACrefatitle{Categorization of natural stimuli by monkeys (Macaca
  mulatta): effects of stimulus set size and modification of
  exemplars.}{Categorization of natural stimuli by monkeys (macaca mulatta):
  effects of stimulus set size and modification of exemplars.}\BBCQ{}
\newblock
\APACjournalVolNumPages{Journal of Experimental Psychology: Animal Behavior
  Processes}{13}{2}{136}.
\PrintBackRefs{\CurrentBib}

\bibitem[\protect\citeauthoryear{%
Simonyan%
\ \BBA{} Zisserman%
}{%
Simonyan%
\ \BBA{} Zisserman%
}{%
{\protect\APACyear{2014}}%
}]{%
simonyan2014very}%
\APACinsertmetastar{%
simonyan2014very}%
Simonyan, K.%
\BCBT{}\ \BBA{} Zisserman, A.%
%
\unskip\
\newblock
\APACrefYearMonthDay{2014}{}{}.
\newblock
\BBOQ{}\APACrefatitle{Very deep convolutional networks for large-scale image
  recognition}{Very deep convolutional networks for large-scale image
  recognition}.\BBCQ{}
\newblock
\APACjournalVolNumPages{arXiv preprint arXiv:1409.1556}{}{}{}.
\PrintBackRefs{\CurrentBib}

\bibitem[\protect\citeauthoryear{%
Stabinger%
, Peer%
\BCBL{}\ \BBA{} Rodríguez-Sánchez%
}{%
Stabinger%
\ \protect\BOthers{.}}{%
{\protect\APACyear{2021}}%
}]{%
stabinger2021nonlocal}%
\APACinsertmetastar{%
stabinger2021nonlocal}%
Stabinger, S.%
, Peer, D.%
\BCBL{}\ \BBA{} Rodríguez-Sánchez, A.%
%
\unskip\
\newblock
\APACrefYearMonthDay{2021}{}{}.
\newblock
\BBOQ{}\APACrefatitle{Arguments for the Unsuitability of Convolutional Neural
  Networks for Non--Local Tasks}{Arguments for the unsuitability of
  convolutional neural networks for non--local tasks}.\BBCQ{}
\newblock
\APACjournalVolNumPages{arXiv preprint arXiv:2102.11944}{}{}{}.
\PrintBackRefs{\CurrentBib}

\bibitem[\protect\citeauthoryear{%
Stabinger%
, Rodr{\'\i}guez-S{\'a}nchez%
\BCBL{}\ \BBA{} Piater%
}{%
Stabinger%
\ \protect\BOthers{.}}{%
{\protect\APACyear{2016}}%
{\protect\APACexlab{{\protect\BCnt{1}}}}}]{%
stabinger201625}%
\APACinsertmetastar{%
stabinger201625}%
Stabinger, S.%
, Rodr{\'\i}guez-S{\'a}nchez, A.%
\BCBL{}\ \BBA{} Piater, J.%
%
\unskip\
\newblock
\APACrefYearMonthDay{2016{\protect\BCnt{1}}}{}{}.
\newblock
\BBOQ{}\APACrefatitle{25 years of CNNs: Can we compare to human abstraction
  capabilities?}{25 years of cnns: Can we compare to human abstraction
  capabilities?}\BBCQ{}
\newblock
\BIn{} \APACrefbtitle{International Conference on Artificial Neural
  Networks}{International conference on artificial neural networks}\ (\BPGS\
  380--387).
\PrintBackRefs{\CurrentBib}

\bibitem[\protect\citeauthoryear{%
Stabinger%
, Rodr{\'\i}guez-S{\'a}nchez%
\BCBL{}\ \BBA{} Piater%
}{%
Stabinger%
\ \protect\BOthers{.}}{%
{\protect\APACyear{2016}}%
{\protect\APACexlab{{\protect\BCnt{2}}}}}]{%
stabinger2016learning}%
\APACinsertmetastar{%
stabinger2016learning}%
Stabinger, S.%
, Rodr{\'\i}guez-S{\'a}nchez, A.%
\BCBL{}\ \BBA{} Piater, J.%
%
\unskip\
\newblock
\APACrefYearMonthDay{2016{\protect\BCnt{2}}}{}{}.
\newblock
\BBOQ{}\APACrefatitle{Learning abstract classes using deep learning}{Learning
  abstract classes using deep learning}.\BBCQ{}
\newblock
\BIn{} \APACrefbtitle{Proceedings of the 9th EAI International Conference on
  Bio-inspired Information and Communications Technologies (formerly
  BIONETICS)}{Proceedings of the 9th eai international conference on
  bio-inspired information and communications technologies (formerly
  bionetics)}\ (\BPGS\ 524--528).
\PrintBackRefs{\CurrentBib}

\bibitem[\protect\citeauthoryear{%
Stabinger%
\ \BBA{} Rodr{\'\i}guez-S{\'a}nchez%
}{%
Stabinger%
\ \BBA{} Rodr{\'\i}guez-S{\'a}nchez%
}{%
{\protect\APACyear{2017}}%
}]{%
stabinger2017evaluation}%
\APACinsertmetastar{%
stabinger2017evaluation}%
Stabinger, S.%
\BCBT{}\ \BBA{} Rodr{\'\i}guez-S{\'a}nchez, A\BPBI J.%
%
\unskip\
\newblock
\APACrefYearMonthDay{2017}{}{}.
\newblock
\BBOQ{}\APACrefatitle{Evaluation of Deep Learning on an Abstract Image
  Classification Dataset.}{Evaluation of deep learning on an abstract image
  classification dataset.}\BBCQ{}
\newblock
\BIn{} \APACrefbtitle{ICCV Workshops}{Iccv workshops}\ (\BPGS\ 2767--2772).
\PrintBackRefs{\CurrentBib}

\bibitem[\protect\citeauthoryear{%
Szegedy%
\ \protect\BOthers{.}}{%
Szegedy%
\ \protect\BOthers{.}}{%
{\protect\APACyear{2014}}%
}]{%
szegedy2014going}%
\APACinsertmetastar{%
szegedy2014going}%
Szegedy, C.%
, Liu, W.%
, Jia, Y.%
, Sermanet, P.%
, Reed, S.%
, Anguelov, D.%
\BCBL{}\ \BOthersPeriod{.}%
\unskip\
\newblock
\APACrefYearMonthDay{2014}{}{}.
\newblock
\BBOQ{}\APACrefatitle{Going deeper with convolutions}{Going deeper with
  convolutions}.\BBCQ{}
\newblock
\APACjournalVolNumPages{arXiv preprint arXiv:1409.4842}{}{}{}.
\PrintBackRefs{\CurrentBib}

\bibitem[\protect\citeauthoryear{%
Teney%
, Anderson%
, He%
\BCBL{}\ \BBA{} Hengel%
}{%
Teney%
\ \protect\BOthers{.}}{%
{\protect\APACyear{2018}}%
}]{%
teney2018tips}%
\APACinsertmetastar{%
teney2018tips}%
Teney, D.%
, Anderson, P.%
, He, X.%
\BCBL{}\ \BBA{} Hengel, A. van~den.%
%
\unskip\
\newblock
\APACrefYearMonthDay{2018}{}{}.
\newblock
\BBOQ{}\APACrefatitle{Tips and tricks for visual question answering: Learnings
  from the 2017 challenge}{Tips and tricks for visual question answering:
  Learnings from the 2017 challenge}.\BBCQ{}
\newblock
\BIn{} \APACrefbtitle{Proceedings of the IEEE Conference on Computer Vision and
  Pattern Recognition}{Proceedings of the ieee conference on computer vision
  and pattern recognition}\ (\BPGS\ 4223--4232).
\PrintBackRefs{\CurrentBib}

\bibitem[\protect\citeauthoryear{%
Teney%
\ \protect\BOthers{.}}{%
Teney%
\ \protect\BOthers{.}}{%
{\protect\APACyear{2019}}%
}]{%
teney2019v}%
\APACinsertmetastar{%
teney2019v}%
Teney, D.%
, Wang, P.%
, Cao, J.%
, Liu, L.%
, Shen, C.%
\BCBL{}\ \BBA{} Hengel, A\BPBI v\BPBI d.%
%
\unskip\
\newblock
\APACrefYearMonthDay{2019}{}{}.
\newblock
\BBOQ{}\APACrefatitle{V-PROM: A Benchmark for Visual Reasoning Using Visual
  Progressive Matrices}{V-prom: A benchmark for visual reasoning using visual
  progressive matrices}.\BBCQ{}
\newblock
\APACjournalVolNumPages{arXiv preprint arXiv:1907.12271}{}{}{}.
\PrintBackRefs{\CurrentBib}

\bibitem[\protect\citeauthoryear{%
Vaswani%
\ \protect\BOthers{.}}{%
Vaswani%
\ \protect\BOthers{.}}{%
{\protect\APACyear{2017}}%
}]{%
attention_is_all_you_need}%
\APACinsertmetastar{%
attention_is_all_you_need}%
Vaswani, A.%
, Shazeer, N.%
, Parmar, N.%
, Uszkoreit, J.%
, Jones, L.%
, Gomez, A\BPBI N.%
\BCBL{}\ \BOthersPeriod{.}%
\unskip\
\newblock
\APACrefYearMonthDay{2017}{}{}.
\newblock
\BBOQ{}\APACrefatitle{Attention is all you need}{Attention is all you
  need}.\BBCQ{}
\newblock
\BIn{} \APACrefbtitle{Advances in neural information processing
  systems}{Advances in neural information processing systems}\ (\BPGS\
  5998--6008).
\PrintBackRefs{\CurrentBib}

\bibitem[\protect\citeauthoryear{%
Vogels%
}{%
Vogels%
}{%
{\protect\APACyear{1999}}%
}]{%
vogels1999categorization}%
\APACinsertmetastar{%
vogels1999categorization}%
Vogels, R.%
%
\unskip\
\newblock
\APACrefYearMonthDay{1999}{}{}.
\newblock
\BBOQ{}\APACrefatitle{Categorization of complex visual images by rhesus
  monkeys. Part 1: behavioural study}{Categorization of complex visual images
  by rhesus monkeys. part 1: behavioural study}.\BBCQ{}
\newblock
\APACjournalVolNumPages{European Journal of Neuroscience}{11}{4}{1223--1238}.
\PrintBackRefs{\CurrentBib}

\bibitem[\protect\citeauthoryear{%
Vonk%
\ \BBA{} MacDonald%
}{%
Vonk%
\ \BBA{} MacDonald%
}{%
{\protect\APACyear{2002}}%
}]{%
vonk2002natural}%
\APACinsertmetastar{%
vonk2002natural}%
Vonk, J.%
\BCBT{}\ \BBA{} MacDonald, S\BPBI E.%
%
\unskip\
\newblock
\APACrefYearMonthDay{2002}{}{}.
\newblock
\BBOQ{}\APACrefatitle{Natural concepts in a juvenile gorilla (Gorilla gorilla
  gorilla) at three levels of abstraction}{Natural concepts in a juvenile
  gorilla (gorilla gorilla gorilla) at three levels of abstraction}.\BBCQ{}
\newblock
\APACjournalVolNumPages{Journal of the experimental analysis of
  behavior}{78}{3}{315--332}.
\PrintBackRefs{\CurrentBib}

\bibitem[\protect\citeauthoryear{%
Vonk%
\ \BBA{} MacDonald%
}{%
Vonk%
\ \BBA{} MacDonald%
}{%
{\protect\APACyear{2004}}%
}]{%
vonk2004levels}%
\APACinsertmetastar{%
vonk2004levels}%
Vonk, J.%
\BCBT{}\ \BBA{} MacDonald, S\BPBI E.%
%
\unskip\
\newblock
\APACrefYearMonthDay{2004}{}{}.
\newblock
\BBOQ{}\APACrefatitle{Levels of abstraction in orangutan (Pongo abelii)
  categorization.}{Levels of abstraction in orangutan (pongo abelii)
  categorization.}\BBCQ{}
\newblock
\APACjournalVolNumPages{Journal of Comparative Psychology}{118}{1}{3}.
\PrintBackRefs{\CurrentBib}

\bibitem[\protect\citeauthoryear{%
Wang%
\ \BBA{} Su%
}{%
Wang%
\ \BBA{} Su%
}{%
{\protect\APACyear{2015}}%
}]{%
wang2015automatic}%
\APACinsertmetastar{%
wang2015automatic}%
Wang, K.%
\BCBT{}\ \BBA{} Su, Z.%
%
\unskip\
\newblock
\APACrefYearMonthDay{2015}{}{}.
\newblock
\BBOQ{}\APACrefatitle{Automatic generation of raven’s progressive
  matrices}{Automatic generation of raven’s progressive matrices}.\BBCQ{}
\newblock
\BIn{} \APACrefbtitle{Twenty-Fourth International Joint Conference on
  Artificial Intelligence.}{Twenty-fourth international joint conference on
  artificial intelligence.}
\PrintBackRefs{\CurrentBib}

\bibitem[\protect\citeauthoryear{%
E.~Wasserman%
, DeVolder%
\BCBL{}\ \BBA{} Coppage%
}{%
E.~Wasserman%
\ \protect\BOthers{.}}{%
{\protect\APACyear{1992}}%
}]{%
wasserman1992non}%
\APACinsertmetastar{%
wasserman1992non}%
Wasserman, E.%
, DeVolder, C.%
\BCBL{}\ \BBA{} Coppage, D.%
%
\unskip\
\newblock
\APACrefYearMonthDay{1992}{}{}.
\newblock
\BBOQ{}\APACrefatitle{Non-similarity-based conceptualization in pigeons via
  secondary or mediated generalization}{Non-similarity-based conceptualization
  in pigeons via secondary or mediated generalization}.\BBCQ{}
\newblock
\APACjournalVolNumPages{Psychological Science}{3}{6}{374--379}.
\PrintBackRefs{\CurrentBib}

\bibitem[\protect\citeauthoryear{%
E\BPBI A.~Wasserman%
, Castro%
\BCBL{}\ \BBA{} Freeman%
}{%
E\BPBI A.~Wasserman%
\ \protect\BOthers{.}}{%
{\protect\APACyear{2012}}%
}]{%
wasserman2012same}%
\APACinsertmetastar{%
wasserman2012same}%
Wasserman, E\BPBI A.%
, Castro, L.%
\BCBL{}\ \BBA{} Freeman, J\BPBI H.%
%
\unskip\
\newblock
\APACrefYearMonthDay{2012}{}{}.
\newblock
\BBOQ{}\APACrefatitle{Same--different categorization in rats}{Same--different
  categorization in rats}.\BBCQ{}
\newblock
\APACjournalVolNumPages{Learning \& Memory}{19}{4}{142--145}.
\PrintBackRefs{\CurrentBib}

\bibitem[\protect\citeauthoryear{%
Wright%
\ \BBA{} Katz%
}{%
Wright%
\ \BBA{} Katz%
}{%
{\protect\APACyear{2006}}%
}]{%
wright2006mechanisms}%
\APACinsertmetastar{%
wright2006mechanisms}%
Wright, A\BPBI A.%
\BCBT{}\ \BBA{} Katz, J\BPBI S.%
%
\unskip\
\newblock
\APACrefYearMonthDay{2006}{}{}.
\newblock
\BBOQ{}\APACrefatitle{Mechanisms of same/different concept learning in primates
  and avians}{Mechanisms of same/different concept learning in primates and
  avians}.\BBCQ{}
\newblock
\APACjournalVolNumPages{Behavioural Processes}{72}{3}{234--254}.
\PrintBackRefs{\CurrentBib}

\bibitem[\protect\citeauthoryear{%
Wu%
\ \protect\BOthers{.}}{%
Wu%
\ \protect\BOthers{.}}{%
{\protect\APACyear{2017}}%
}]{%
wu2017visual}%
\APACinsertmetastar{%
wu2017visual}%
Wu, Q.%
, Teney, D.%
, Wang, P.%
, Shen, C.%
, Dick, A.%
\BCBL{}\ \BBA{} Hengel, A. van~den.%
%
\unskip\
\newblock
\APACrefYearMonthDay{2017}{}{}.
\newblock
\BBOQ{}\APACrefatitle{Visual question answering: A survey of methods and
  datasets}{Visual question answering: A survey of methods and
  datasets}.\BBCQ{}
\newblock
\APACjournalVolNumPages{Computer Vision and Image
  Understanding}{163}{}{21--40}.
\PrintBackRefs{\CurrentBib}

\bibitem[\protect\citeauthoryear{%
Xu%
\ \protect\BOthers{.}}{%
Xu%
\ \protect\BOthers{.}}{%
{\protect\APACyear{2015}}%
}]{%
xu2015show}%
\APACinsertmetastar{%
xu2015show}%
Xu, K.%
, Ba, J.%
, Kiros, R.%
, Cho, K.%
, Courville, A.%
, Salakhudinov, R.%
\BCBL{}\ \BOthersPeriod{.}%
\unskip\
\newblock
\APACrefYearMonthDay{2015}{}{}.
\newblock
\BBOQ{}\APACrefatitle{Show, attend and tell: Neural image caption generation
  with visual attention}{Show, attend and tell: Neural image caption generation
  with visual attention}.\BBCQ{}
\newblock
\BIn{} \APACrefbtitle{International conference on machine
  learning}{International conference on machine learning}\ (\BPGS\ 2048--2057).
\PrintBackRefs{\CurrentBib}

\bibitem[\protect\citeauthoryear{%
Yun%
, Bohn%
\BCBL{}\ \BBA{} Ling%
}{%
Yun%
\ \protect\BOthers{.}}{%
{\protect\APACyear{2020}}%
}]{%
yun2020deeper}%
\APACinsertmetastar{%
yun2020deeper}%
Yun, X.%
, Bohn, T.%
\BCBL{}\ \BBA{} Ling, C.%
%
\unskip\
\newblock
\APACrefYearMonthDay{2020}{}{}.
\newblock
\BBOQ{}\APACrefatitle{A Deeper Look at Bongard Problems}{A deeper look at
  bongard problems}.\BBCQ{}
\newblock
\BIn{} \APACrefbtitle{Canadian Conference on Artificial Intelligence}{Canadian
  conference on artificial intelligence}\ (\BPGS\ 528--539).
\PrintBackRefs{\CurrentBib}

\bibitem[\protect\citeauthoryear{%
Zaremba%
, Sutskever%
\BCBL{}\ \BBA{} Vinyals%
}{%
Zaremba%
\ \protect\BOthers{.}}{%
{\protect\APACyear{2014}}%
}]{%
zaremba2014recurrent}%
\APACinsertmetastar{%
zaremba2014recurrent}%
Zaremba, W.%
, Sutskever, I.%
\BCBL{}\ \BBA{} Vinyals, O.%
%
\unskip\
\newblock
\APACrefYearMonthDay{2014}{}{}.
\newblock
\BBOQ{}\APACrefatitle{Recurrent neural network regularization}{Recurrent neural
  network regularization}.\BBCQ{}
\newblock
\APACjournalVolNumPages{arXiv preprint arXiv:1409.2329}{}{}{}.
\PrintBackRefs{\CurrentBib}

\bibitem[\protect\citeauthoryear{%
T.~Zentall%
\ \BBA{} Hogan%
}{%
T.~Zentall%
\ \BBA{} Hogan%
}{%
{\protect\APACyear{1976}}%
}]{%
zentall1976}%
\APACinsertmetastar{%
zentall1976}%
Zentall, T.%
\BCBT{}\ \BBA{} Hogan, D.%
%
\unskip\
\newblock
\APACrefYearMonthDay{1976}{02}{}.
\newblock
\BBOQ{}\APACrefatitle{Pigeons Can Learn Identity or Difference, or
  Both}{Pigeons can learn identity or difference, or both}.\BBCQ{}
\newblock
\APACjournalVolNumPages{Science (New York, N.Y.)}{191}{}{408-9}.
\PrintBackRefs{\CurrentBib}

\bibitem[\protect\citeauthoryear{%
T\BPBI R.~Zentall%
, Galizio%
\BCBL{}\ \BBA{} Critchfield%
}{%
T\BPBI R.~Zentall%
\ \protect\BOthers{.}}{%
{\protect\APACyear{2002}}%
}]{%
zentall2002categorization}%
\APACinsertmetastar{%
zentall2002categorization}%
Zentall, T\BPBI R.%
, Galizio, M.%
\BCBL{}\ \BBA{} Critchfield, T\BPBI S.%
%
\unskip\
\newblock
\APACrefYearMonthDay{2002}{}{}.
\newblock
\BBOQ{}\APACrefatitle{Categorization, concept learning, and behavior analysis:
  An introduction}{Categorization, concept learning, and behavior analysis: An
  introduction}.\BBCQ{}
\newblock
\APACjournalVolNumPages{Journal of the experimental analysis of
  behavior}{78}{3}{237--248}.
\PrintBackRefs{\CurrentBib}

\bibitem[\protect\citeauthoryear{%
Zhang%
, Bengio%
, Hardt%
, Recht%
\BCBL{}\ \BBA{} Vinyals%
}{%
Zhang%
\ \protect\BOthers{.}}{%
{\protect\APACyear{2016}}%
}]{%
rethink-generalization}%
\APACinsertmetastar{%
rethink-generalization}%
Zhang, C.%
, Bengio, S.%
, Hardt, M.%
, Recht, B.%
\BCBL{}\ \BBA{} Vinyals, O.%
%
\unskip\
\newblock
\APACrefYearMonthDay{2016}{}{}.
\newblock
\BBOQ{}\APACrefatitle{Understanding deep learning requires rethinking
  generalization}{Understanding deep learning requires rethinking
  generalization}.\BBCQ{}
\newblock
\APACjournalVolNumPages{arXiv preprint arXiv:1611.03530}{}{}{}.
\PrintBackRefs{\CurrentBib}

\end{thebibliography}

\end{document}